\documentclass{article}

\usepackage[preprint]{corl_2024} % Uncomment for pre-prints (e.g., arxiv); This is like ``final'', but will remove the CORL footnote.
\usepackage{graphicx}
\usepackage{enumitem}
\usepackage{booktabs}
\usepackage{multicol}
\usepackage{multirow}
\usepackage{appendix}
\usepackage{listings}
\usepackage{amsmath}
\usepackage{wrapfig}
\usepackage{courier}
\usepackage{pifont}
\usepackage{algorithm}
\usepackage[noend]{algorithmic}
\usepackage[breakable]{tcolorbox}

\usepackage{xspace}

\newcommand{\ie}{\textit{\textrm{i.e.}}}
\newcommand{\eg}{\textit{\textrm{e.g.}}}

\newcommand{\cmark}{\ding{51}}%
\newcommand{\xmark}{\ding{55}}%

\def\graphlong{Open Scene Graph}
\def\graphshort{OSG}
\def\enginelong{OSG Mapper}
\def\engineshort{OSG Mapper}
\def\navsys{\textit{OpenSearch}}
\def\edgecontains{``\texttt{contains}''}
\def\edgeconnects{``\texttt{connects to}''}
\def\edgenear{``\texttt{is near}''}

\def\ourspr{OS}
\def\oursprgt{OS-GT}
\def\hm{HM3D}
\def\hmfull{HM3D-Semantics v0.1}
\def\gibson{Gibson}

\def\vo{v^{o}}

\def\vp{v^{p}}
\def\vi{v^{i}}

\def\objnode{$\vo{}$}
\def\objnodelabel{$\vo_{label}$}
\def\objnodeimage{$\vo_{img}$}
\def\objnodeid{$\vo_{id}$}
\def\objnodetext{$\vo_{desc}$}

\def\placenode{$\vp{}$}
\def\placenodeclass{$\vp_{cls}$}
\def\placenodelabel{$\vp_{label}$}
\def\placenodeid{$\vp_{id}$}

\def\absinodelabel{$\vi_{label}$}
\def\absinodeid{$\vi_{id}$}

\def\centroidthresh{$\beta_{pix}$}
\def\bboxthresh{$\beta_{IOU}$}

\def\ooclass{\textsc{Cls}}
\def\ooattr{\textsc{Att}}
\def\oodom{\textsc{Dom}}
\def\oorel{\textsc{Rel}}
\def\ooclassabs{\textsc{Cls}_{abs}}
\def\ooattrabs{\textsc{Att}_{abs}}
\def\oodomabs{\textsc{Dom}_{abs}}
\def\oorelabs{\textsc{Rel}_{abs}}
\def\oostate{\mathcal{S}}
\def\ooaction{\mathcal{A}}
\def\ootrans{\mathcal{T}}
\def\ooreward{\mathcal{R}}
\def\oohoriz{\gamma}
\def\ooobs{\Omega}
\def\ooobsmodel{O}
\def\ooobjs{\mathcal{O}}
\def\ooedges{\mathcal{E}}

\def\weblink{\urllink[pre = \bgroup\bf, post = \egroup]}

\usepackage{etoolbox}
\makeatletter
\patchcmd{\hyper@makecurrent}{%
    \ifx\Hy@param\Hy@chapterstring
        \let\Hy@param\Hy@chapapp
    \fi
}{%
    \iftoggle{inappendix}{%true-branch
        \@checkappendixparam{chapter}%
        \@checkappendixparam{section}%
        \@checkappendixparam{subsection}%
        \@checkappendixparam{subsubsection}%
        \@checkappendixparam{paragraph}%
        \@checkappendixparam{subparagraph}%
    }{}%
}{}{\errmessage{failed to patch}}

\newcommand*{\@checkappendixparam}[1]{%
    \def\@checkappendixparamtmp{#1}%
    \ifx\Hy@param\@checkappendixparamtmp
        \let\Hy@param\Hy@appendixstring
    \fi
}
\makeatletter

\newtoggle{inappendix}
\togglefalse{inappendix}

\apptocmd{\appendix}{\toggletrue{inappendix}}{}{\errmessage{failed to patch}}
\apptocmd{\subappendices}{\toggletrue{inappendix}}{}{\errmessage{failed to patch}}
\title{\graphlong{}s for Open World \\Object-Goal Navigation}

\author{
  Joel Loo$^{*}$, Zhanxin Wu$^{*}$, David Hsu\\
  Smart Systems Institute, National University of Singapore\\
  \texttt{\{joell, zhanxinwu, dyhsu\}@comp.nus.edu.sg} \\
  \href{https://open-scene-graphs.github.io}{https://open-scene-graphs.github.io}
}

\begin{document}
\maketitle
\def\thefootnote{*}\footnotetext{Equal contribution.}

%===============================================================================

\begin{abstract}
    How can we build robots for open-world semantic navigation tasks, like searching for target objects in novel scenes? While foundation models have the rich knowledge and generalisation needed for these tasks, a \textit{suitable scene representation} is needed to connect them into a complete robot system. We address this with \textit{\graphlong{}s} (\graphshort{}s), a topo-semantic representation that retains and organises open-set scene information for these models, and has a structure that can be configured for different environment types. We integrate foundation models and \graphshort{}s into the \textit{\navsys{}} system for Open World Object-Goal Navigation, which is capable of searching for open-set objects specified in natural language, while generalising zero-shot across diverse environments and embodiments. Our \graphshort{}s enhance reasoning with Large Language Models (LLM), enabling robust object-goal navigation outperforming existing LLM approaches. Through simulation and real-world experiments, we validate \textit{\navsys{}}'s generalisation across varied environments, robots and novel instructions.

\end{abstract}

% Two or three meaningful keywords should be added here

%===============================================================================

\section{Introduction}

  \begin{figure}[!t]
  % \setlength{\linewidth}{\textwidth}
  % \setlength{\hsize}{\textwidth}
  % \centering
  \includegraphics[width=\linewidth]{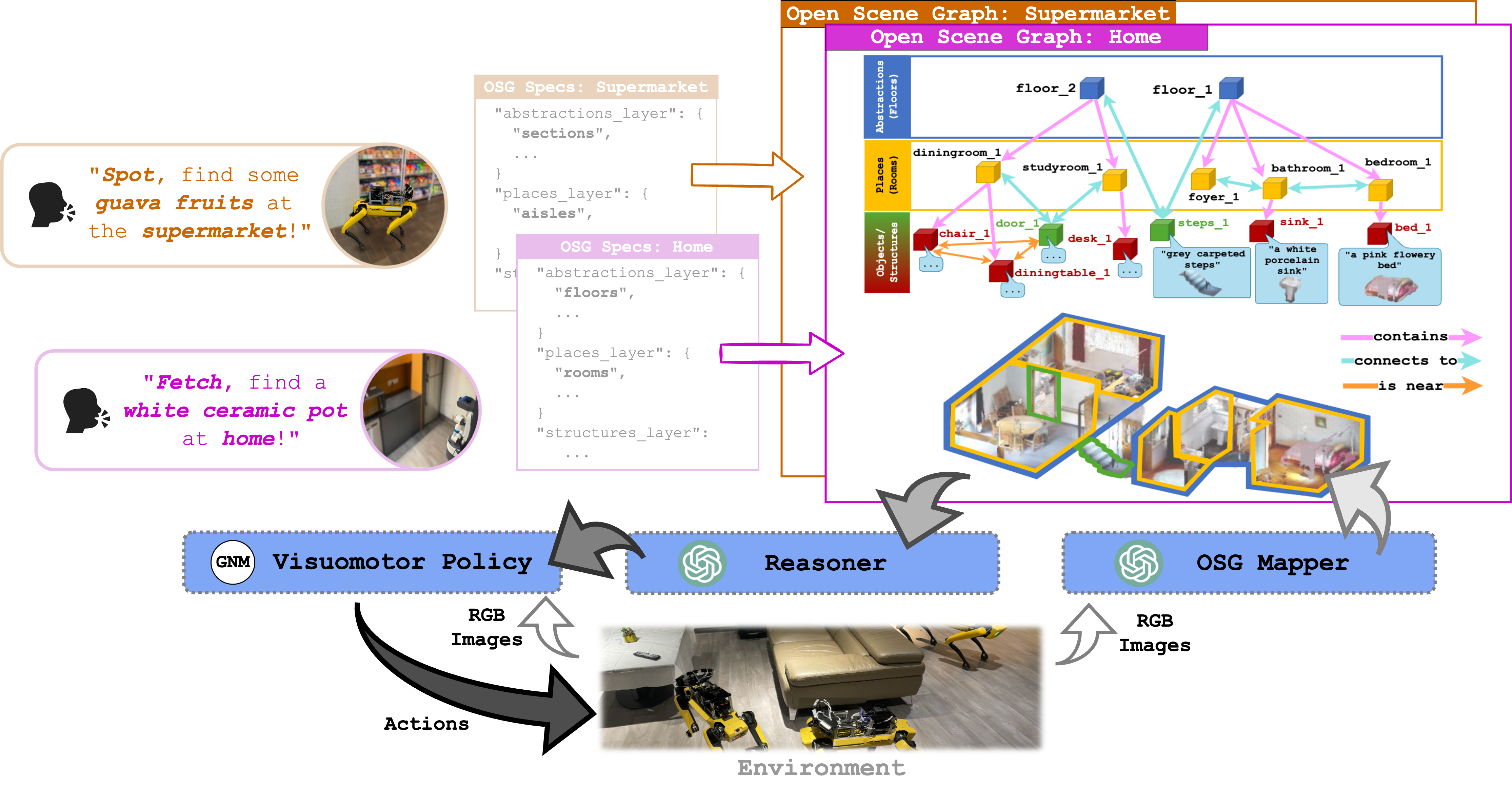}
  \caption{\textbf{\textit{\graphlong{}s} enable Open World ObjectNav systems.} We design the \textit{\navsys{}} system for open-vocabulary ObjectNav across environments and embodiments. Our \graphshort{} acts as a suitable semantic scene memory for the foundation models that provide \navsys{}'s semantic understanding and generalisation. Notably, \graphshort{}s facilitate generalisation over diverse environments since their structures can be dynamically configured to best represent the current environment type.}
  \label{fig:splash}
  \end{figure}

Consider being asked to find a wine-bottle in an unfamiliar house. We might draw on our common-sense to guess at likely places for wine - \eg{} kitchen or even a wine cellar - then reason about their physical locations and explore the house to find them. \textit{Semantic navigation} tasks that go beyond geometric scene understanding, to also require semantic reasoning and priors, are commonplace in daily life~\cite{yokoyama2024vlfm}. Robots not only need to be capable of such tasks but also perform them in realistic open-world settings: they may need to find unusual objects, to search a diverse range of environments, or to be robust to changes in sensing and dynamics owing to hardware upgrades or wear.

In this work, we study \textit{object-goal navigation} (ObjectNav)~\cite{habitatchallenge2022} extended to the \textit{open world}, wherein the robot is tasked with navigating to specified object categories in novel unmapped scenes, while further needing to generalise zero-shot across diverse environments, embodiments and to novel instructions. Recent foundation models offer a means to tackle this Open World ObjectNav (OWON) task: Large Language Models (LLMs) are capable of semantic reasoning and have rich knowledge of human environments~\cite{zhou2023navgpt, chen2024mapgpt, yu2023l3mvn, shah2023lfg}, while Visual Foundation Models (VFMs)~\cite{kirillov2023segany, liu2023grounding, li2023blip2, minderer2022owlvit} and General Navigation Models (GNMs)~\cite{shah2023vint, sridhar2023nomad} show strong cross-embodiment and cross-environment performance. Composing a robot system fully from foundation models thus offers a means to achieve the level of semantic understanding and generalisation needed in OWON.

Though foundation models are powerful building blocks, a \textit{suitable scene representation} is needed to complete the system and serve as the working memory connecting the models. We seek a representation that can be built purely with foundation models, and provides the structured and open-ended scene understanding OWON requires. Specifically we desire that scene information is organised in a semantically meaningful structure to facilitate reasoning with LLMs~\cite{rana2023sayplan, zheng2024stepback}, and that the representation be designed to effectively capture scenes across diverse environment types. 

To bridge this gap, we present the \textit{\graphlong{}} (\graphshort{}), a hierarchical, topo-semantic representation to capture open-set scene information about \textit{objects} and \textit{spatial regions} in images and text. Typical scene graphs have fixed structures tailored to specific environment types~\cite{armeni2019scenegraph, rosinol2021kimera} like offices, making it hard to build and use them zero-shot elsewhere. We imbue \graphshort{}s with a configurable structure, which can be instantiated for a given environment type (\eg{} homes or supermarkets) from a \textit{meta-structure}. The meta-structure is defined to capture elements of a valid \graphshort{} structure that stay constant \textit{over environment types}. Algorithms to build/use \graphshort{}s can also be specified as abstract program sketches defined in terms of the meta-structure: these work by prompting foundation models, and exploit their open-vocabulary abilities to handle new \graphshort{} structure instances.

% We give \graphshort{}s a configurable structure, by defining a \textit{meta-structure} from which they can be instantiated for a given environment type, \eg{} homes, supermarkets. The meta-structure captures the elements of a valid \graphshort{} structure that stay constant \textit{across environment types}.

We show the feasibility of building an autonomous robot system for OWON with \textit{\navsys{}}, a system composed fully from foundation models that uses an \graphshort{} as its core representation. We find that \graphshort{}s improve LLMs' reasoning on ObjectNav tasks, enabling us to outperform by a wide margin existing LLM approaches using less informative and structured representations. Through simulation and real world experiments, we show that \navsys{} enables effective open-vocabulary ObjectNav that generalises zero-shot over diverse indoor environments and robot embodiments.

% built by FMs and effectively used by them for OWON -> latter means that they must generalise across environments + contain information in form usable by FMs + structured for LLM comprehension

%===============================================================================

% \section{Citations}
% \label{sec:citations}

% 	Citations can be made using either \textbackslash citep\{\} or \textbackslash citet\{\}, depending from the appropriateness. To avoid the citation moving to the next line, it is often a good practice to replace the space before with a tilde (\~{}) character.
% 	Example 1: ``CoRL is the best conference ever~\citep{Gauss1857}.''
% 	Example 2: ``\citet{Lagrange1788} proved, both theoretically and numerically, that CoRL is the best conference ever.''
	
%===============================================================================

\section{Open World Object-Goal Navigation (OWON) Task}
\label{sec:task}

ObjectNav~\cite{habitatchallenge2022} is the task of searching for an instance of a specified object category in a novel unmapped indoor environment. We extend it to an open world setting in OWON: we seek a robot system capable of handling open-set object queries specified in natural language, across a diverse range of environments and robot embodiments. In OWON, a robot's input is an egocentric forward-facing RGB feed, and it outputs linear and angular velocity commands for actuation. OWON relaxes several assumptions made in ObjectNav. For sensing, we only assume visual inputs and allow camera model/mounting height to vary. For embodiments, we allow any robot that can be commanded with linear/angular velocities. We assume unavailability of metric information, \eg{} pose estimates: these can require metric SLAM, limiting generalisation over sensors, robots and scenes~\cite{Cadena2016PastPA}. Overall, this setting encourages solutions that \textit{generalise across embodiments and environments}.

% ObjectNav~\cite{habitatchallenge2022} is the task of searching for an instance of a specified object category in a novel unmapped indoor environment. We extend it to an open world setting in OWON: we seek a robot system capable of handling open-set object queries specified in natural language, across a diverse range of environments and robot embodiments. In OWON, a robot's input is an egocentric forward-facing RGB feed, and it outputs linear and angular velocity commands for actuation. The OWON setup differs from ObjectNav in 3 ways. Firstly, we relax assumptions on sensing: we only assume visual inputs, allowing camera model and mounting height to vary. Secondly, we relax assumptions on embodiment: different robots are allowed if they can be commanded with linear/ angular velocities. Thirdly, we assume unavailability of metric information like pose estimates: these can depend on metric SLAM systems with limited generalisation over sensors, robots and scenes~\cite{Cadena2016PastPA}. Overall, this setting encourages solutions that \textit{generalise across embodiments and environments}.

\section{Approach}
\label{sec:approach}

\subsection{Preliminaries}
\label{sec:approach_prelim}

We model the object search problem in OWON as an extension of Object-Oriented POMDPs (OO-POMDPs)~\cite{diuk2008oomdp, wandzel2019mos}. OO-POMDPs use object state abstractions to reason about objects and their relations. An OO-POMDP is factored into a \textit{schema} $\langle \ooclass{}, \ooattr{}, \oodom{}, \oorel{} \rangle$ that abstractly defines the state over a range of domains, and domain-specific \textit{instances} $\langle \oostate{}, \ooaction{}, \ootrans{}, \ooreward{}, \ooobs{}, \ooobsmodel{}, \oohoriz{} \rangle$. Concretely, state $\oostate{}$ comprises a set of objects $\ooobjs{}$, where $o_i\in\ooobjs{}$ is an instance of \textit{class} $c\in\ooclass{}$, and a set of relations between objects $\ooedges{}$, where $e(o_j, o_k)\in\ooedges{}$ links $o_j, o_k$ with edge type $e\in\oorel{}$. Each $o_i\in\ooobjs{}$ has attributes $\ooattr{}(c) = \{a_1,...,a_m\}$, where each attribute $a$ takes a value in range $\oodom(a)$.

While OO-POMDPs are usually applied to object-centric tasks like multi-object search~\cite{wandzel2019mos}, existing formulations often do not reason over the scene's \textit{spatial structure}, or even \textit{non-goal objects}~\cite{wandzel2019mos, zheng2020mos, zheng2023genmos}. Such contextual information is vital for updating our belief on a target object's location~\cite{grinvald2019volumetric, chen2023semutil}: \eg{} a dining-room can hint at an adjoining kitchen containing the target wine-bottle, while wine-glasses can suggest that wine is nearby. Thus, we expand OO-POMDPs to capture all observed objects and spatial regions in our state, updating them from egocentric partial observations. 

Formally, we define an OWON OO-POMDP with state $\oostate{}$ comprising the agent's state $s_a$, the set of observed spatial regions $s_r$, and the set of observed objects $s_o$. $\ooclass{}$ specifies that $r_i\in s_r$, $o_j\in s_o$ are instances of semantically meaningful, open-set region/object categories. $\oorel{}$ is extended to define spatial connectivity and containment relations among regions/objects, in addition to spatial proximity object relations in usual OO-POMDPs. Observations $\ooobs{}$ are egocentric RGB images, while action space $\ooaction{}$ contains the high-level primitives \textsc{MoveToObject}($o$) and \textsc{MoveToRegion}($r$). 

We posit that foundation models can help to approximate solutions for OWON OO-POMDPs. Firstly, VFMs' provide the open-set visual understanding of scene objects and spatial structure needed to reason about OWON. Secondly, LLMs' rich semantic knowledge helps to model OWON's complex belief dynamics. We can use LLMs to implicitly maintain and update a belief in the target object's location by prompting them with observed scene information, reducing the OO-POMDP to an OO-Belief MDP and abstracting away the challenge of specifying an observation model.

% We hypothesise that foundation models help approximate solutions to OWON OO-POMDPs in 3 ways. Firstly, LLMs' rich semantic priors allow them to model OWON's complex belief dynamics. We can guide LLMs to implicitly maintain and update a belief in the target object's location by prompting them with observed semantic scene information, reducing the OO-POMDP to an OO-Belief MDP, and abstracting away the challenge of specifying an observation model. Secondly, LLMs' semantic ``reasoning'' abilities enable them to act as effective policies that map the implicit belief to effective actions for search. Thirdly, VFMs' open-set visual understanding provides the rich semantic information about scene objects and spatial structure needed to reason about OWON.

% - LLMs implicitly maintain and update belief in target object 
% - LLMs effectively capture belief dynamics and can maintain and update belief in target object locations
% - Simplifies overall process to belief MDP, no need to explicitly define observation model

\subsection{\navsys{} system overview}

  \begin{figure}[!t]
  \includegraphics[width=\linewidth, height=134pt]{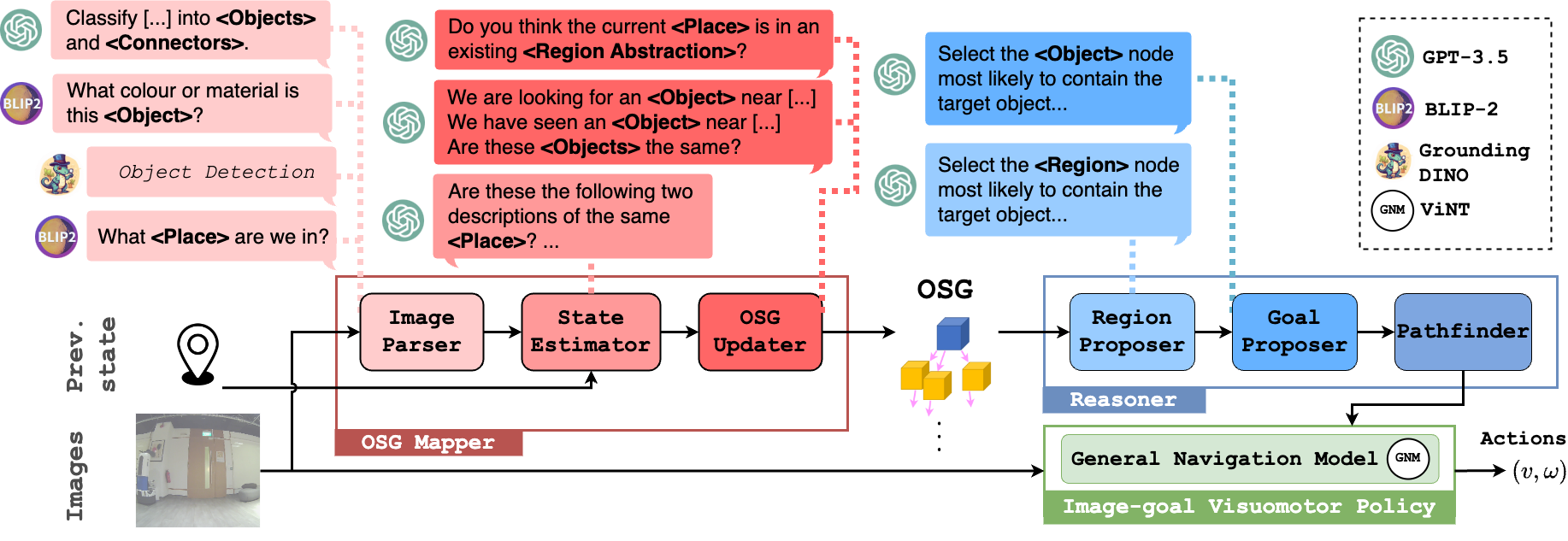}
  % \caption{\textbf{System architecture of \navsys{}.}}
  \caption{\textbf{\navsys{} system overview.} \textit{\engineshort{}} and \textit{Reasoner} build/reason with the \graphshort{} using templated prompts. These prompts are grounded with concepts specified in an \graphshort{} spec.}
  \label{fig:system}
  \end{figure}

We present \navsys{} which uses foundation models zero-shot with \graphshort{}s for OWON. Since OWON OO-POMDPs need object/region information and relational structure, scene graphs are ideal. We introduce \graphshort{}s, which provide \textit{rich, structured, open-set scene information} for downstream reasoning. These generalise over embodiments and environments with a topo-semantic design to reduce reliance on metric information, and a configurable structure. This is achieved with \graphshort{}s' meta-structure, an \textit{abstract schema} broadly defining valid \graphshort{} structure elements across indoor environments, which can be instantiated into an \graphshort{} structure for specific environment types.

\navsys{} (\autoref{fig:system}) comprises 3 modules composed from foundation models: the \textit{\engineshort{}} for online \graphshort{} building, the \textit{Reasoner} to maintain a belief in the target's location and propose search actions, and the \textit{Controller} for navigation. The first two are abstract programs sketching out a series of prompts. Given a user-specified \graphshort{} structure, the sketches' prompts can be grounded in the environment type. All modules are built from 4 types of foundation models: an LLM, a GNM, and VFMs for Visual Question Answering (VQA) and Open-set Object Detection.

% We present \navsys{}, which uses foundation models zero-shot with \graphshort{}s for OWON. The need for object/region-centric information and relational structure in the OWON OO-POMDP's state motivates the use of scene graphs like \graphshort{}s. \graphshort{}s are especially designed for open-world settings, with configurable structures enabling cross-environment generalisation, and a topo-semantic design that minimises reliance on metric information to enable cross-embodiment generalisation.

% We use foundation models in the roles described above, composing the modules that make up \navsys{} from them: the \textit{\engineshort{}} for online \graphshort{} building, the \textit{Reasoner} to maintain a belief in the target's location and propose search actions, and the \textit{Controller} for navigation. We show how these modules can be constructed from 4 existing types of foundation models: an LLM, a GNM, and VFMs for Visual Question Answering (VQA) and Open-set Object Detection.

\subsection{\graphlong{}s}
\label{sec:approach_osg}

An \textit{\graphlong{}} (\graphshort{}) is a structured representation of scene objects and spatial regions over multiple levels of abstraction. Formally, it is a heterogeneous, simple directed graph organised into $N$ layers. We abstractly specify \graphshort{}s with a \textit{meta-structure}, \ie{} the abstract schema $\langle \ooclassabs{}, \ooattrabs{}, \oodomabs{}, \oorelabs{} \rangle$. For specific environment types (\eg{} homes, supermarkets, malls) this is grounded into the schema $\langle \ooclass{}, \ooattr{}, \oodom{}, \oorel{} \rangle$, \ie{} an \graphshort{} specification describing the \graphshort{}'s structure, as in \autoref{fig:splash}. Examples and further details are given in \autoref{app:osg_details}.

\subsubsection{Abstract node/edge types}

We define in $\ooclassabs{}$, 4 abstract types of nodes. \textbf{(N1) Objects.} Static scene elements occupying spatially localised regions: \eg{} tables, chairs etc. They provide semantic cues for reasoning, and act as navigation goals and localisation landmarks - specifically, we distinguish Objects/Places with \textit{object features}, aggregated lists of nearby objects. \textbf{(N2) Places.} The smallest semantically meaningful spatial regions in a scene, \eg{} \textit{rooms} in a home or \textit{aisles} in a supermarket, which typically cannot be subdivided in a consistent, meaningful way. \textbf{(N3) Region Abstractions.} Spatial abstractions encompassing multiple smaller spatial regions: \eg{} a \textit{floor} containing multiple \textit{rooms}, or a supermarket \textit{section} that is a cluster of \textit{aisles}. Such hierarchical abstractions can aid reasoning, especially with LLMs~\cite{zheng2024stepback}. \textbf{(N4) Connectors.} These are specific types of Objects that are also structural scene elements connecting spatial regions: \eg{} \textit{doors} connecting rooms, \textit{steps} connecting floors.

We define in $\oorelabs{}$, 3 abstract types of directed edges. \textbf{(E1) Spatial Proximity.} Outgoing \edgenear{} edges approximately capture closeness between Object/Connector nodes, and specify which neighbours to use in the source node's \textit{object features}. \textbf{(E2) Spatial Connectivity.} \edgeconnects{} edges specify reachability between nodes, and can connect Place, Region Abstraction and Structure nodes. \textbf{(E3) Hierarchy.} \edgecontains{} edges specify spatial containment relations among regions/objects, and implicitly define a tree-like hierarchy of spatial abstractions on the \graphshort{}.

% Describe form of the OSG
\subsubsection{Meta-structure of an \graphshort{}}

\textbf{Layer 1: Objects.} \textit{(Required).} Leaf nodes representing distinct open-set objects, labelled in language. They can have outgoing \edgenear{} edges to close Object/Connector nodes. Object nodes have attributes describing its appearance in the form of open-vocabulary text and images.

\textbf{Layer 2: Connectors.} \textit{(Optional).} Connectors have similar properties as Objects, but also can have \edgeconnects{} edges to spatial region nodes. An \graphshort{} spec should specify the various classes of Connectors that may be found in a given type of environment, \eg{} doors, stairs etc. Since not all environments have Connectors (\eg{} open-plan offices) they are optional.

\textbf{Layer 3: Places.} \textit{(Required).} An \graphshort{} spec should define the semantically meaningful classes of Places in a given environment type, \eg{} \textit{rooms}, \textit{corridors} in homes. If possible, each Place node is given an open-set semantic text label that is an instance of its Place class, \eg{} \textit{kitchen}, \textit{dining-room} for rooms. They can have \edgeconnects{} edges to adjacent Place/Connector nodes, and \edgecontains{} edges to Object nodes (which also define a Place's object features).

\textbf{Layers 4-$N$: Region Abstractions.} \textit{(Optional).} Each layer defines a different spatial abstraction and has similar properties to Places, with the additional condition that nodes in the $i$th layer may have \edgecontains{} edges only to nodes in the $(i-1)$th layer.

\subsection{\enginelong{} module}
\label{sec:approach_mapper}
The \enginelong{} is an abstract sketch of a map-building algorithm, which takes input RGB images, processes them into text with VFMs, then prompts LLMs with the text to update the \graphshort{}. With an \graphshort{} spec, it is instantiated into an executable mapping routine which builds the \graphshort{} based on the structure defined in the spec. It comprises \textbf{(1)} an image parser to extract semantic information from RGB observations; \textbf{(2)} a state estimator; \textbf{(3)} an \graphshort{} updater to integrate parsed information into the \graphshort{}. We provide further details on algorithms, prompts and hyperparameters in \autoref{app:mapper_details}.

\textbf{Image Parser.} We extract image and open-vocabulary textual information about Places, Objects and Connectors from RGB images. We obtain the current Place's semantic label by prompting the VQA model to identify the Place type (\textit{room}), then to suggest the instance label based on it (\textit{kitchen}). An open-set object detector identifies and labels Objects and Connectors in the scene, and the VQA model provides textual descriptions of each based on image crops. Spatial proximity between Objects and Connectors is heuristically determined by thresholding pixel distances.

\textbf{State Estimator.} The state estimator uses an LLM to match extracted textual observations from the Image Parser with the \graphshort{} to identify the robot's state $s_a\in s_r\cup\{\phi\}$. Initially, the estimator identifies candidate Places within the \graphshort{} by using the LLM to evaluate the semantic similarity between the observed Place label and the Place node labels in the \graphshort{}. It then performs place recognition by comparing the observed object features with the candidate Places' stored features, while focusing on the similarity of descriptions, particularly for larger objects.

% \textbf{\graphshort{} Updater.} The \graphshort{} Updater incrementally integrates observations from the Image Parser into the \graphshort{} using an LLM. When the robot is at an existing Place node, it updates the Place's contents, and otherwise adds new spatial region nodes. In the first scenario, the LLM is prompted to use object features for data association, to determine whether observations should update existing Object/Connector nodes or add new ones. In the second scenario, new Object, Connector, and Place nodes are added directly from observations, and updates are propagated upwards to layers $4-N$. Geometric Region Abstractions, \eg{} \textit{floors} connected by \textit{stairs}, can be programmatically updated based on spatial connectivity. For layers involving semantic abstractions, \eg{} \textit{sections} of a supermarket, the LLM's semantic understanding is used to decide if the current spatial region belongs to an existing Region Abstraction node or if a new node is needed.

\textbf{\graphshort{} Updater.} The \graphshort{} Updater incrementally integrates observations from the Image Parser into the \graphshort{} using an LLM. At an existing Place node, it updates the node's contents by using the LLM to associate observations with existing leaf nodes, based on object features. Otherwise it adds new Place and leaf nodes from observations, then propagates updates up the layers. Some upper layers represent Region Abstractions defined solely by scene geometry, which can be updated based on spatial connectivity (\eg{} \textit{floors} can be inferred from the presence of \textit{stairs}). Other Region Abstractions are defined by both spatial and semantic cues (\eg{} \textit{sections} in a supermarket), requiring the use of the LLM to decide if the current location lies in an existing or new node.

% When at an existing Place node, it updates the Place's contents; otherwise, it adds new spatial region nodes. In the former, the LLM uses object features for data association to decide if observations should update existing leaf nodes or add new ones. Otherwise, new Place/leaf nodes are added from observations, and updates propagated up the layers. 

\subsection{Reasoner module}
\label{sec:approach_reasoner}

% The \textit{Reasoner} is an abstract algorithm sketch which uses the scene object and spatial structure information in the \graphshort{} to reason and propose the next action in the object search. As described in \autoref{sec:approach_prelim}, it uses LLMs to generate a belief in the target object's location then select an action. By passing in a \graphshort{} spec for a particular environment type, we can instantiate from it an executable routine that takes an input JSON string of the \graphshort{} and selects an object $o\in s_o$ to move toward. Since objects are an \graphshort{}'s most granular elements, we only implement and use the \textsc{MoveToObject} primitive in $\ooaction{}$. The routine first provides a prompt encoding the \graphshort{} and the target object, to guide the LLM to implicitly update its belief in the object's location. Based on the implicit belief, the \textit{Region Proposer} guides the LLM to select promising regions to search. It does so by iterating from Layers $N$ to 3 and prompting the LLM to select a child node at each layer. Given a selected Place, the \textit{Goal Proposer} then guides the LLM to identify a specific Object/Connector leaf node contained or adjacent to the Place to search. We use few-shot prompting and guide the LLM to output its reasoning~\cite{shah2023lfg} to improve performance (\autoref{app:reasoner_details}). Finally, given an Object/Connector goal node, the \textit{Pathfinder} module searches for a feasible path over the \graphshort{} with Dijkstra, tracks the robot's progress and triggers replanning if needed.

The \textit{Reasoner} is an abstract algorithm sketch that uses scene object and spatial structure information from \graphshort{}s to reason. It employs LLMs to update the belief in the target object's location and select the next search action. It is grounded into an executable routine with an \graphshort{} spec: this takes a JSON string of the \graphshort{} and outputs an object $o\in s_o$ to navigate to with \textsc{MoveToObject}. We only take object-centric actions since objects are \graphshort{}s' most granular elements. The routine prompts an LLM with the \graphshort{}, target object and action history, guiding it to update its implicit belief. Based on this belief, the \textit{Region Proposer} guides the LLM to choose a promising Place to search by iteratively choosing a child node at each spatial region layer from $N$ to 3. The \textit{Goal Proposer} then selects an Object/Connector leaf node of the chosen Place, to search. We use few-shot prompting, and have the LLM detail its reasoning to enhance performance~\cite{shah2023lfg} (\autoref{app:reasoner_details}). Finally, the \textit{Pathfinder} finds a feasible path in the \graphshort{} with Dijkstra, tracks the robot's progress, and triggers replanning if needed.

\subsection{Controller module}
\label{sec:approach_controller}

We implement \textsc{MoveToObject} using image-goal visuomotor policies, which have enabled cross-embodiment navigation models~\cite{shah2023vint, sridhar2023nomad} and can operate on the existing image information stored in \graphshort{}s. Concretely, we use the ViNT~\cite{shah2023vint} GNM zero-shot for visual navigation toward target objects specified by image crops saved in the \graphshort{}. Implementation details are given in \autoref{app:controller_details}.

% Image-goal visuomotor policies have been shown to be good substrates for building cross-embodiment navigation models~\cite{shah2023vint, sridhar2023nomad}, and are compatible with the many current foundation models that operate on non-metric, visual inputs. The latter makes them well-suited for our system since the other subsystems and the \graphshort{} purely use the modalities of vision and language. Specifically, \navsys{} uses the ViNT GNM~\cite{shah2023vint} zero-shot, and commands it with image crops of a nearby object to navigate to. We find that the raw temporal distance-to-goal values predicted by ViNT vary with scenario and embodiment, and are unreliable for precisely stopping the robot. Instead of directly thresholding on these, we use a heuristic stopping criterion motivated by the observation that the temporal distance values tend to drop as the robot nears the goal, before sharply rising again due to a significant appearance change of the goal stemming from the closer viewpoint. This leads us to use this ``minimum'' as a stopping criterion: we set a threshold on the maximum positive change from the minimum recorded distance to trigger the robot to halt.

\section{Experimental Results}
\label{sec:result}
Our experiments answer the questions: \textbf{Q1.} Is LLM planning with \graphshort{}s effective for ObjectNav over diverse environments? \textbf{Q2.} How do \graphshort{}s enhance exploration with LLM planners? \textbf{Q3.} How effective are object features for data association?  \textbf{Q4.} How well can we build \graphshort{}s over diverse environments? \textbf{Q5.} How well does \navsys{} perform across diverse open-vocabulary goals? \textbf{Q6.} How well does \navsys{} perform in the real world, across different robot embodiments?

\subsection{Simulation experiments}
\label{sec:sim_exp_setup}

\begin{table}[t]
    \centering
    \begin{minipage}[b]{0.4\textwidth}
        \centering
        \footnotesize
        \caption{\textbf{ObjectNav performance of LLM baselines in HM3D.}}
        \label{tab:llm_ablations}
            \begin{tabular}{@{\extracolsep{4pt}}cccc@{}}
                \toprule
                \textbf{Method} & \textbf{SR} ($\uparrow$) & \textbf{SPL} ($\uparrow$) & \textbf{DTG}($\downarrow$) \\
                \midrule
                LGX-GT~\cite{dorbala2024lgx} & 0.275 & 0.080 & 5.078  \\
                LFG-GT~\cite{shah2023lfg} & 0.675 & \textbf{0.389} & 2.411 \\
                \oursprgt{} &\textbf{0.775} & 0.380 & \textbf{1.702} \\
                \midrule
                \ourspr{} & 0.693 & 0.283 & 2.338 \\
                \bottomrule
            \end{tabular}
    \end{minipage}
    \hfill
    \begin{minipage}[b]{0.5\textwidth}
    \centering
    \scriptsize
    \caption{\textbf{Comparing with ObjectNav approaches in Gibson.} TF / NM describe training-free/non-metric approaches.}
    \label{tab:gibson_results}
        \begin{tabular}{@{\extracolsep{4pt}}ccccc@{}}
            \toprule
            \textbf{Method} & \textbf{SR} ($\uparrow$) & \textbf{SPL} ($\uparrow$) & \textbf{DTG} ($\downarrow$) & \textbf{TF} / \textbf{NM} \\
            \midrule
            SemExp~\cite{chaplot2020semexp} & 0.657 & 0.339 & 1.474 & \xmark / \xmark  \\
            PONI~\cite{ramakrishnan2022poni} & \textbf{0.736} & \textbf{0.410} & \textbf{1.250} & \xmark / \xmark \\
            \midrule
            FBE~\cite{yamauchi1997fbe} & 0.641 & 0.283 & 1.780 & \cmark / \xmark\\
            SemUtil~\cite{chen2023semutil} & 0.693 & \textbf{0.405} & \textbf{1.488} & \cmark / \xmark \\
            \midrule
            \ourspr{} &\textbf{0.734} & 0.386 & 1.722 & \cmark / \cmark\\
            \bottomrule
        \end{tabular}
    \end{minipage}
\end{table}

\begin{figure}[!t]
  \includegraphics[width=\linewidth, height=153pt]{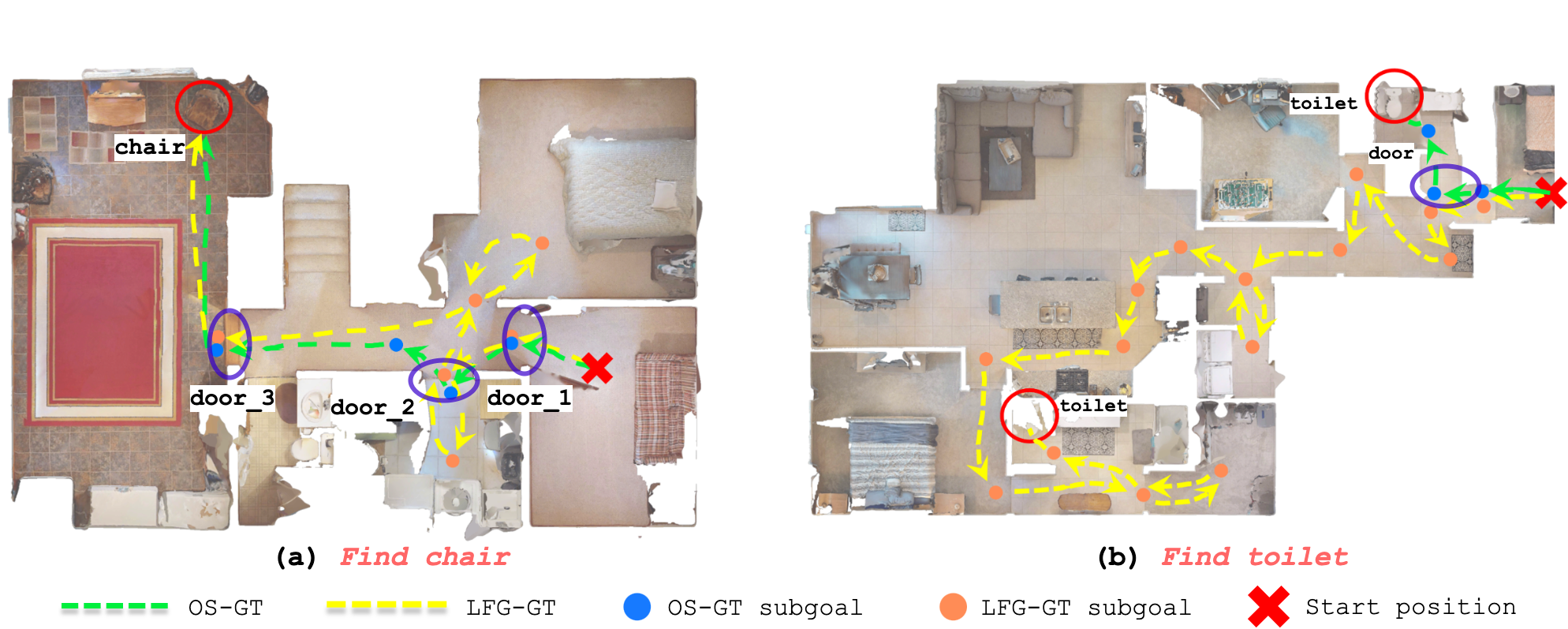}
  \caption{\textbf{Value of topo-semantic information on scene structure to LLM-based ObjectNav.} In both cases, there are few objects near the robot's starting position and hence sparse object cues. (a): LFG defaults to metric search, while OS is guided by its region-level spatial understanding to efficiently sweep through the rooms. (b): OS recognises it is starting in a bedroom and searches nearby Connectors, while LFG lacks region/structure understanding and searches in the wrong direction.}
  \label{fig:analysis}
\end{figure}

\begin{table}[t]
    \centering
    \begin{minipage}[b]{0.35\textwidth}
        \centering
        \caption{\textbf{Accuracy of data association with object features.} We test on 
        \textit{recognising} a scene element from different views, and \textit{distinguishing} different elements.}
        \label{tab:obj_feat}
        \scriptsize
            \begin{tabular}{@{\extracolsep{4pt}}cccc@{}}
                \toprule
                \textbf{Node} & \textbf{Recognise} & \multicolumn{2}{c}{\textbf{Distinguish}} \\
                \cmidrule{3-4}
                & & \textit{Same type} & \textit{Diff. type} \\
                \midrule
                Objects & 0.85 & 0.58 & - \\
                Connectors & 0.75 & 0.83 & - \\
                Places & 0.88 & 0.80 & 0.94 \\
                \bottomrule
            \end{tabular}
    \end{minipage}
    \hfill
    \begin{minipage}[b]{0.55\textwidth}
    \centering
    \caption{\textbf{Evaluation of scene graph quality.} We measure \textbf{Pr}ecision and \textbf{Re}call in the estimation of nodes and outgoing edges for \textit{floors} (Region Abstractions), \textit{rooms} (Places), \textit{doors} (Connectors)}
    \label{tab:scene_graph_quality}
    \scriptsize
        \begin{tabular}{@{\extracolsep{4pt}}ccccccc@{}}
            \toprule
            \textbf{}&\multicolumn{2}{c}{Nodes}&\multicolumn{2}{c}{\edgecontains{}}&\multicolumn{2}{c}{\edgeconnects{}} \\
            \cmidrule{2-3}\cmidrule{4-5}\cmidrule{6-7}
            \textbf{} & \textbf{Pr} & \textbf{Re} & \textbf{Pr} & \textbf{Re} & \textbf{Pr} & \textbf{Re} \\
            \midrule
            \textit{Floors}  & 1.000 & 1.000 & 0.889 & 0.914 & 1.000 & 1.000   \\
            \textit{Rooms} & 0.846 & 0.880 & - & - & 0.771 & 0.831 \\
            \textit{Doors} &0.845 & 0.800 & - & - & 0.783 & 0.857 \\
            \bottomrule
        \end{tabular}
    \end{minipage}
\end{table}

We evaluate \textbf{Q1-5} in simulation. Further details on our setup and baselines are given in \autoref{app:sim_exp_details}.

\textbf{Datasets and metrics.} We use the \gibson{}~\cite{xiazamirhe2018gibsonenv} and \hmfull{}~\cite{ramakrishnan2021hm3d} datasets in the Habitat simulator~\cite{szot2021habitat} and evaluate on the metrics from the Habitat ObjectNav challenge~\cite{habitatchallenge2022}: success rate (SR), success-weighted path length (SPL) and distance-to-goal (DTG) in metres.
    
\textbf{Baselines.} We compare against recent prompt-based LLM ObjectNav approaches, which can be viewed as ablations of \navsys{}. We use the Fast Marching Method~\cite{sethian1996fmm} across all baselines and \navsys{} variants, to isolate performance improvements due to representations and reasoning. Baselines relying on ground-truth semantic annotations are marked with \textbf{GT}. We compare the following: \textbf{(B1)} \textit{LGX-GT}~\cite{dorbala2024lgx}. Prompts LLM with objects in immediate field-of-view, to decide a direction to search in. \textbf{(B2)} \textit{LFG-GT}~\cite{shah2023lfg}. Prompts LLM with list of object clusters from the frontiers in a global metric map, to decide promising frontiers to search. \textbf{(B3)} \textit{\oursprgt{}}. Our \navsys{} system prompts LLMs with both object and region information in the form of an \graphshort{}, to decide on promising places to search. \textbf{(B4)} \textit{\ourspr{}}. An \navsys{} variant using VFMs for perception rather than GT annotations. Apart from LLM baselines, we also evaluate \ourspr{} on Gibson to compare it with state-of-the-art ObjectNav approaches that are (i) trained on ObjectNav datasets (\ie{} SemExp~\cite{chaplot2020semexp}, PONI~\cite{ramakrishnan2022poni}), (ii) training-free (\ie{} FBE~\cite{yamauchi1997fbe}, SemUtil~\cite{chen2023semutil}).

To address \textbf{Q1}, we find that \navsys{} has strong zero-shot performance on ObjectNav across diverse indoor environments. On \gibson{} (\autoref{tab:gibson_results}), \ourspr{} achieves a high success rate competitive with the strongest learned baselines like PONI, despite not having ObjectNav-specific training. We also attain a high success rate on \hm{}, which has a greater diversity of household scenes. Notably, we show stronger performance than other training-free approaches, like the recent SemUtil~\cite{chen2023semutil} which combines classical planning with closed-set semantics, and outperform recent methods using LLMs by a wide margin (\autoref{tab:llm_ablations}). While our topo-semantic approach does not let us optimise for shortest geometric paths, the \navsys{} variants still remain competitive with state-of-the-art baselines in SPL and DTG. In particular, we note that we also outperform by a large margin frontier-based exploration (FBE)~\cite{yamauchi1997fbe}, a purely metric exploration approach. This highlights the value of semantic priors and reasoning for efficient search across diverse environments. % Add in discussion of limitations here?

For \textbf{Q2}, we find that structured representations capturing both \textit{objects} and \textit{spatial regions} enhance ObjectNav performance. From \autoref{tab:llm_ablations}, LFG-GT improves significantly over LGX-GT, highlighting the need for a global scene memory. \oursprgt{} outperforms LFG-GT by additionally retaining information on regions and scene topology in the \graphshort{}, suggesting its value to reasoning for ObjectNav. \autoref{fig:analysis} highlights how the LLM can make use of \graphshort{} structure and topology information (\eg{} Places, Connectors) for efficient search, especially when object information is scarce.

\begin{wraptable}{r}{0.33\textwidth}
    \centering
    \caption{\textbf{Real-world results on OWON.} Success rate on 5 trials.}
    \label{tab:real_world_exp}
    \small
        \begin{tabular}{@{\extracolsep{4pt}}ccc@{}}
            \toprule
            \textbf{Object goal} & \textit{Spot} & \textit{Fetch} \\
            \midrule
            \texttt{guitar} & 4/5 &  4/5 \\
            \texttt{dish washer} & 5/5 &  4/5 \\
            \bottomrule
        \end{tabular}
\end{wraptable}

For \textbf{Q3}, we evaluate the accuracy of associating Objects, Connectors and Places with object features, using sampled views across 10 \hm{} scenes. From \autoref{tab:obj_feat} we see that object features can reliably recognise and distinguish objects/regions. We note that they can struggle to correctly distinguish two distinct objects of similar type, as human environments often have similar objects clustered closely (\eg{} chairs grouped together around a table). Each object's neighbours (and hence features) are too similar, causing aliasing. For \textbf{Q4} we show that the \engineshort{} can map varied environments accurately. From \autoref{tab:scene_graph_quality}, it has good accuracy in predicting an \graphshort{}'s nodes and edges. \autoref{app:scene_graph_results} shows and discusses examples of \graphshort{}s built to represent scenes in both simulation and real-world.

For \textbf{Q5}, we show that \navsys{} handles open-vocabulary object queries well. We validate this with \textit{rare} and \textit{compositional} queries. The former are uncommon, long-tailed object concepts, and we test on examples from the LVIS dataset~\cite{gupta2019lvis} (\eg{} settee). The latter involve objects composed with open-vocabulary descriptors (\eg{} leopard-print fabric chair). We find that \navsys{} generalises over both kinds of queries, and provide quantitative results in \autoref{app:open_vocab}.

\subsection{Real-world experiments}

We evaluate zero-shot generalisation to different robots and novel object queries in an open-plan apartment on two robots: \textbf{(a)} \textit{Spot}, a quadrupedal robot with a $170^\circ$ fisheye RGB camera at $\sim$~0.8m height; \textbf{(b)} \textit{Fetch}, a differential-drive robot with a $79^\circ$ Realsense RGB camera at $\sim$~1.4m height. We test with less common objects not in ObjectNav and present our results in \autoref{tab:real_world_exp} and a supplementary video. The same system and models are used across experiments without parameter tuning. While \navsys{} often succeeds, we note that the GNM's performance on goal-directed navigation in cluttered areas is limited, inducing the observed failures.

\section{Related Work}
\label{sec:related_work}

\textbf{ObjectNav.} Recent works~\cite{wahid2018sac, mousavian2019visualrep, maksymets2021thda, ye2021auxiliary, ramakrishnan2022poni} have made notable progress on benchmarks, yet often require training on curated datasets for ObjectNav or adjacent tasks, leading to limited generalisation~\cite{majumdar2022zson}. Foundation models hold promise to address this issue, given their impressive generalisation and open-world semantic knowledge~\cite{hu2023robotfmsurvey}. A growing body of work explores such models' utility to ObjectNav, primarily to handle open-set object queries~\cite{majumdar2022zson, yu2023l3mvn, dorbala2024lgx}. Motivated by these models' generality, we propose to extend ObjectNav to the more realistic OWON task, that additionally seeks generalisation across instructions, embodiments and environments. We show with \navsys{} that composing foundation models zero-shot offers a feasible solution to OWON.

% - extended task to more realistic open world setting
% - demonstrate a feasible solution for this task, by composing foundation models zero-shot

\textbf{LLMs for reasoning in navigation.} LLMs' apparent ability to perform reasoning~\cite{hu2023robotfmsurvey, huang2022survey} in addition to their generalisation makes them key foundation models in robot systems. They are increasingly applied to task planning~\cite{rana2023sayplan} and semantic reasoning~\cite{shah2023lfg, dorbala2024lgx, zhou2023navgpt, chen2024mapgpt}. Many existing works elicit reasoning from LLMs by prompting them with textual scene information of \textit{objects} alone~\cite{dorbala2024lgx, cai2023pixnav, yu2023l3mvn, shah2023lfg}. This does not consider information on spatial regions, which is also vital for reasoning~\cite{wandzel2019mos, honerkamp2024momallm}. SayPlan~\cite{rana2023sayplan} shows effective LLM reasoning given rich information on \textit{objects} and \textit{regions} captured in pre-built scene graphs. Our work not only provides LLMs with rich, structured, open-set scene information in \graphshort{}s, but also presents an online \graphshort{} mapping method.

\textbf{Scene graphs.} These are topological representations capturing spatial concepts with hierarchies of abstractions~\cite{armeni2019scenegraph, rosinol2021kimera}, which are often applied to task and motion planning~\cite{amiri2022reasoningscenegraphs, ravichandran2022hierarchical, rana2023sayplan, ray2024tampscenegraphs}. Kimera~\cite{rosinol2021kimera} is an archetypal approach: it specifies a fixed graph structure comprising rooms, floors, buildings, and extracts these spatial concepts from a metric-semantic mesh. Recent works extend scene graphs to capture open-vocabulary spatial concepts~\cite{strader2024ontology, werby2024hovsg}. \graphshort{}s are designed for broader generalisation. They not only capture open-set spatial concepts, but organise them in a configurable structure that can be adapted to different environments. \graphshort{} mapping purely uses VFMs and LLMs without relying on metric SLAM, facilitating generalisation across sensors and robots.

%===============================================================================

\section{Conclusion}
\label{sec:conclusion}

Our OWON task extends ObjectNav to require generalisation across instructions, environments and embodiments. We propose the \navsys{} system for OWON, which uses foundation models to reason and generalise. Key to it is our \graphshort{} representation, which captures rich, structured, open-set scene information about \textit{objects} and \textit{regions} that facilitates LLM reasoning. We show that our system generalises zero-shot across open-set language object queries, environments and robots. Future directions include inferring \graphshort{} specs and using \navsys{}'s promptability for lifelong learning.

\textbf{Limitations.} Our system can take 0.5-2 minutes to update the scene graph and propose actions, due to our heavy reliance on LLMs (GPT-3.5). Ongoing work on smaller language models~\cite{mitra2023orca} for local inference holds promise for addressing this. While we do not handle uncertainty presently due to the challenge of obtaining accurate uncertainty estimates from LLMs and VFMs, we note that recent works~\cite{ren2023knowno, ren2024explore} using tools like conformal prediction can offer a way to address this issue.

% - open-set object and region information as input
% - foundation models for open-set map-building, belief update and search policy

% Though \navsys{} possesses strong open-world generalisation, it does so by forgoing standard engineered robot system components like metric-based SLAM and local planners, which are designed to enable robustness and precision through \textit{uncertainty handling} and effective use of \textit{metric information} respectively. We envision that future work on \navsys{} can yield a more capable system with high performance and open-world generalisation.

%===============================================================================

\clearpage
% The acknowledgments are automatically included only in the final and preprint versions of the paper.
% \acknowledgments{If a paper is accepted, the final camera-ready version will (and probably should) include acknowledgments. All acknowledgments go at the end of the paper, including thanks to reviewers who gave useful comments, to colleagues who contributed to the ideas, and to funding agencies and corporate sponsors that provided financial support.}

%===============================================================================

% no \bibliographystyle is required, since the corl style is automatically used.
\bibliography{references}  % .bib

\begin{thebibliography}{50}
\providecommand{\natexlab}[1]{#1}
\providecommand{\url}[1]{\texttt{#1}}
\expandafter\ifx\csname urlstyle\endcsname\relax
  \providecommand{\doi}[1]{doi: #1}\else
  \providecommand{\doi}{doi: \begingroup \urlstyle{rm}\Url}\fi

\bibitem[Yokoyama et~al.(2024)Yokoyama, Ha, Batra, Wang, and Bucher]{yokoyama2024vlfm}
N.~Yokoyama, S.~Ha, D.~Batra, J.~Wang, and B.~Bucher.
\newblock Vlfm: Vision-language frontier maps for zero-shot semantic navigation.
\newblock In \emph{International Conference on Robotics and Automation (ICRA)}, 2024.

\bibitem[Yadav et~al.(2022)Yadav, Ramakrishnan, Turner, Gokaslan, Maksymets, Jain, Ramrakhya, Chang, Clegg, Savva, Undersander, Chaplot, and Batra]{habitatchallenge2022}
K.~Yadav, S.~K. Ramakrishnan, J.~Turner, A.~Gokaslan, O.~Maksymets, R.~Jain, R.~Ramrakhya, A.~X. Chang, A.~Clegg, M.~Savva, E.~Undersander, D.~S. Chaplot, and D.~Batra.
\newblock Habitat challenge 2022.
\newblock \url{https://aihabitat.org/challenge/2022/}, 2022.

\bibitem[Zhou et~al.(2023)Zhou, Hong, and Wu]{zhou2023navgpt}
G.~Zhou, Y.~Hong, and Q.~Wu.
\newblock Navgpt: Explicit reasoning in vision-and-language navigation with large language models, 2023.

\bibitem[Chen et~al.(2024)Chen, Lin, Xu, Chai, Liang, and Wong]{chen2024mapgpt}
J.~Chen, B.~Lin, R.~Xu, Z.~Chai, X.~Liang, and K.-Y.~K. Wong.
\newblock Mapgpt: Map-guided prompting for unified vision-and-language navigation, 2024.

\bibitem[Yu et~al.(2023)Yu, Kasaei, and Cao]{yu2023l3mvn}
B.~Yu, H.~Kasaei, and M.~Cao.
\newblock L3mvn: Leveraging large language models for visual target navigation.
\newblock In \emph{IEEE/RSJ International Conference on Intelligent Robots and Systems}, pages 3554--3560, 10 2023.

\bibitem[Shah et~al.(2023)Shah, Equi, Osi{\'n}ski, Xia, Ichter, and Levine]{shah2023lfg}
D.~Shah, M.~R. Equi, B.~Osi{\'n}ski, F.~Xia, B.~Ichter, and S.~Levine.
\newblock Navigation with large language models: Semantic guesswork as a heuristic for planning.
\newblock In \emph{7th Annual Conference on Robot Learning}, 2023.

\bibitem[Kirillov et~al.(2023)Kirillov, Mintun, Ravi, Mao, Rolland, Gustafson, Xiao, Whitehead, Berg, Lo, Doll{\'a}r, and Girshick]{kirillov2023segany}
A.~Kirillov, E.~Mintun, N.~Ravi, H.~Mao, C.~Rolland, L.~Gustafson, T.~Xiao, S.~Whitehead, A.~C. Berg, W.-Y. Lo, P.~Doll{\'a}r, and R.~Girshick.
\newblock Segment anything.
\newblock \emph{arXiv:2304.02643}, 2023.

\bibitem[Liu et~al.(2023)Liu, Zeng, Ren, Li, Zhang, Yang, Li, Yang, Su, Zhu, et~al.]{liu2023grounding}
S.~Liu, Z.~Zeng, T.~Ren, F.~Li, H.~Zhang, J.~Yang, C.~Li, J.~Yang, H.~Su, J.~Zhu, et~al.
\newblock Grounding dino: Marrying dino with grounded pre-training for open-set object detection.
\newblock \emph{arXiv preprint arXiv:2303.05499}, 2023.

\bibitem[Li et~al.(2023)Li, Li, Savarese, and Hoi]{li2023blip2}
J.~Li, D.~Li, S.~Savarese, and S.~Hoi.
\newblock {BLIP-2:} bootstrapping language-image pre-training with frozen image encoders and large language models.
\newblock In \emph{ICML}, 2023.

\bibitem[Minderer et~al.(2022)Minderer, Gritsenko, Stone, Neumann, Weissenborn, Dosovitskiy, Mahendran, Arnab, Dehghani, Shen, Wang, Zhai, Kipf, and Houlsby]{minderer2022owlvit}
M.~Minderer, A.~Gritsenko, A.~Stone, M.~Neumann, D.~Weissenborn, A.~Dosovitskiy, A.~Mahendran, A.~Arnab, M.~Dehghani, Z.~Shen, X.~Wang, X.~Zhai, T.~Kipf, and N.~Houlsby.
\newblock Simple open-vocabulary object detection.
\newblock In \emph{Computer Vision – ECCV 2022: 17th European Conference, Tel Aviv, Israel, October 23–27, 2022, Proceedings, Part X}, page 728–755, Berlin, Heidelberg, 2022. Springer-Verlag.

\bibitem[Shah et~al.(2023)Shah, Sridhar, Dashora, Stachowicz, Black, Hirose, and Levine]{shah2023vint}
D.~Shah, A.~Sridhar, N.~Dashora, K.~Stachowicz, K.~Black, N.~Hirose, and S.~Levine.
\newblock Vi{NT}: A foundation model for visual navigation.
\newblock In \emph{7th Annual Conference on Robot Learning}, 2023.

\bibitem[Sridhar et~al.(2023)Sridhar, Shah, Glossop, and Levine]{sridhar2023nomad}
A.~Sridhar, D.~Shah, C.~Glossop, and S.~Levine.
\newblock {NoMaD: Goal Masked Diffusion Policies for Navigation and Exploration}.
\newblock \emph{arXiv pre-print}, 2023.

\bibitem[Rana et~al.(2023)Rana, Haviland, Garg, Abou-Chakra, Reid, and Suenderhauf]{rana2023sayplan}
K.~Rana, J.~Haviland, S.~Garg, J.~Abou-Chakra, I.~Reid, and N.~Suenderhauf.
\newblock Sayplan: Grounding large language models using 3d scene graphs for scalable robot task planning.
\newblock In \emph{7th Annual Conference on Robot Learning}, 2023.

\bibitem[Zheng et~al.(2024)Zheng, Mishra, Chen, Cheng, Chi, Le, and Zhou]{zheng2024stepback}
H.~S. Zheng, S.~Mishra, X.~Chen, H.-T. Cheng, E.~H. Chi, Q.~V. Le, and D.~Zhou.
\newblock Step-back prompting enables reasoning via abstraction in large language models.
\newblock In \emph{The Twelfth International Conference on Learning Representations}, 2024.

\bibitem[Armeni et~al.(2019)Armeni, He, Gwak, Zamir, Fischer, Malik, and Savarese]{armeni2019scenegraph}
I.~Armeni, Z.-Y. He, J.~Gwak, A.~R. Zamir, M.~Fischer, J.~Malik, and S.~Savarese.
\newblock 3d scene graph: A structure for unified semantics, 3d space, and camera.
\newblock In \emph{Proceedings of the IEEE International Conference on Computer Vision}, 2019.

\bibitem[Rosinol et~al.(2021)Rosinol, Violette, Abate, Hughes, Chang, Shi, Gupta, and Carlone]{rosinol2021kimera}
A.~Rosinol, A.~Violette, M.~Abate, N.~Hughes, Y.~Chang, J.~Shi, A.~Gupta, and L.~Carlone.
\newblock Kimera: From slam to spatial perception with 3d dynamic scene graphs.
\newblock \emph{The International Journal of Robotics Research}, 40\penalty0 (12-14):\penalty0 1510--1546, 2021.

\bibitem[Cadena et~al.(2016)Cadena, Carlone, Carrillo, Latif, Scaramuzza, Neira, Reid, and Leonard]{Cadena2016PastPA}
C.~Cadena, L.~Carlone, H.~Carrillo, Y.~Latif, D.~Scaramuzza, J.~Neira, I.~D. Reid, and J.~J. Leonard.
\newblock Past, present, and future of simultaneous localization and mapping: Toward the robust-perception age.
\newblock \emph{IEEE Transactions on Robotics}, 32:\penalty0 1309--1332, 2016.

\bibitem[Diuk et~al.(2008)Diuk, Cohen, and Littman]{diuk2008oomdp}
C.~Diuk, A.~Cohen, and M.~L. Littman.
\newblock An object-oriented representation for efficient reinforcement learning.
\newblock In \emph{Proceedings of the 25th International Conference on Machine Learning}, 2008.

\bibitem[Wandzel et~al.(2019)Wandzel, Oh, Fishman, Kumar, Wong, and Tellex]{wandzel2019mos}
A.~Wandzel, Y.~Oh, M.~Fishman, N.~Kumar, L.~L. Wong, and S.~Tellex.
\newblock Multi-object search using object-oriented pomdps.
\newblock In \emph{2019 International Conference on Robotics and Automation (ICRA)}, 2019.

\bibitem[Zheng et~al.(2020)Zheng, Sung, Konidaris, and Tellex]{zheng2020mos}
K.~Zheng, Y.~Sung, G.~D. Konidaris, and S.~Tellex.
\newblock Multi-resolution pomdp planning for multi-object search in 3d.
\newblock \emph{2021 IEEE/RSJ International Conference on Intelligent Robots and Systems (IROS)}, 2020.

\bibitem[Zheng et~al.(2023)Zheng, Paul, and Tellex]{zheng2023genmos}
K.~Zheng, A.~Paul, and S.~Tellex.
\newblock A system for generalized 3d multi-object search.
\newblock In \emph{2023 IEEE International Conference on Robotics and Automation (ICRA)}, pages 1638--1644, 2023.

\bibitem[{Grinvald} et~al.(2019){Grinvald}, {Furrer}, {Novkovic}, {Chung}, {Cadena}, {Siegwart}, and {Nieto}]{grinvald2019volumetric}
M.~{Grinvald}, F.~{Furrer}, T.~{Novkovic}, J.~J. {Chung}, C.~{Cadena}, R.~{Siegwart}, and J.~{Nieto}.
\newblock {Volumetric Instance-Aware Semantic Mapping and 3D Object Discovery}.
\newblock \emph{IEEE Robotics and Automation Letters}, July 2019.

\bibitem[Chen et~al.(2023)Chen, Li, Kumar, Ghanem, and Yu]{chen2023semutil}
J.~Chen, G.~Li, S.~Kumar, B.~Ghanem, and F.~Yu.
\newblock {How To Not Train Your Dragon: Training-free Embodied Object Goal Navigation with Semantic Frontiers}.
\newblock In \emph{Proceedings of Robotics: Science and Systems}, Daegu, Republic of Korea, July 2023.

\bibitem[Dorbala et~al.(2024)Dorbala, Mullen, and Manocha]{dorbala2024lgx}
V.~S. Dorbala, J.~F. Mullen, and D.~Manocha.
\newblock Can an embodied agent find your “cat-shaped mug”? llm-based zero-shot object navigation.
\newblock \emph{IEEE Robotics and Automation Letters}, 9\penalty0 (5):\penalty0 4083--4090, 2024.

\bibitem[Chaplot et~al.(2020)Chaplot, Gandhi, Gupta, and Salakhutdinov]{chaplot2020semexp}
D.~S. Chaplot, D.~Gandhi, A.~Gupta, and R.~Salakhutdinov.
\newblock Object goal navigation using goal-oriented semantic exploration.
\newblock In \emph{In Neural Information Processing Systems (NeurIPS)}, 2020.

\bibitem[Ramakrishnan et~al.(2022)Ramakrishnan, Chaplot, Al-Halah, Malik, and Grauman]{ramakrishnan2022poni}
S.~K. Ramakrishnan, D.~S. Chaplot, Z.~Al-Halah, J.~Malik, and K.~Grauman.
\newblock Poni: Potential functions for objectgoal navigation with interaction-free learning.
\newblock In \emph{Computer Vision and Pattern Recognition (CVPR), 2022 IEEE Conference on}. IEEE, 2022.

\bibitem[Yamauchi(1997)]{yamauchi1997fbe}
B.~Yamauchi.
\newblock A frontier-based approach for autonomous exploration.
\newblock In \emph{Proceedings 1997 IEEE International Symposium on Computational Intelligence in Robotics and Automation CIRA'97. 'Towards New Computational Principles for Robotics and Automation'}, 1997.

\bibitem[Xia et~al.(2018)Xia, R.~Zamir, He, Sax, Malik, and Savarese]{xiazamirhe2018gibsonenv}
F.~Xia, A.~R.~Zamir, Z.-Y. He, A.~Sax, J.~Malik, and S.~Savarese.
\newblock Gibson env: real-world perception for embodied agents.
\newblock In \emph{Computer Vision and Pattern Recognition (CVPR), 2018 IEEE Conference on}. IEEE, 2018.

\bibitem[Ramakrishnan et~al.(2021)Ramakrishnan, Gokaslan, Wijmans, Maksymets, Clegg, Turner, Undersander, Galuba, Westbury, Chang, Savva, Zhao, and Batra]{ramakrishnan2021hm3d}
S.~K. Ramakrishnan, A.~Gokaslan, E.~Wijmans, O.~Maksymets, A.~Clegg, J.~M. Turner, E.~Undersander, W.~Galuba, A.~Westbury, A.~X. Chang, M.~Savva, Y.~Zhao, and D.~Batra.
\newblock Habitat-matterport 3d dataset ({HM}3d): 1000 large-scale 3d environments for embodied {AI}.
\newblock In \emph{Thirty-fifth Conference on Neural Information Processing Systems Datasets and Benchmarks Track}, 2021.

\bibitem[Szot et~al.(2021)Szot, Clegg, Undersander, Wijmans, Zhao, Turner, Maestre, Mukadam, Chaplot, Maksymets, Gokaslan, Vondrus, Dharur, Meier, Galuba, Chang, Kira, Koltun, Malik, Savva, and Batra]{szot2021habitat}
A.~Szot, A.~Clegg, E.~Undersander, E.~Wijmans, Y.~Zhao, J.~Turner, N.~Maestre, M.~Mukadam, D.~Chaplot, O.~Maksymets, A.~Gokaslan, V.~Vondrus, S.~Dharur, F.~Meier, W.~Galuba, A.~Chang, Z.~Kira, V.~Koltun, J.~Malik, M.~Savva, and D.~Batra.
\newblock Habitat 2.0: Training home assistants to rearrange their habitat.
\newblock In \emph{Advances in Neural Information Processing Systems (NeurIPS)}, 2021.

\bibitem[Sethian(1996)]{sethian1996fmm}
J.~A. Sethian.
\newblock A fast marching level set method for monotonically advancing fronts.
\newblock \emph{Proceedings of the National Academy of Sciences}, 93\penalty0 (4):\penalty0 1591--1595, 1996.

\bibitem[Gupta et~al.(2019)Gupta, Dollar, and Girshick]{gupta2019lvis}
A.~Gupta, P.~Dollar, and R.~Girshick.
\newblock {LVIS}: A dataset for large vocabulary instance segmentation.
\newblock In \emph{Proceedings of the {IEEE} Conference on Computer Vision and Pattern Recognition}, 2019.

\bibitem[Wahid et~al.(2021)Wahid, Stone, Chen, Ichter, and Toshev]{wahid2018sac}
A.~Wahid, A.~Stone, K.~Chen, B.~Ichter, and A.~Toshev.
\newblock Learning object-conditioned exploration using distributed soft actor critic.
\newblock In \emph{Proceedings of the 2020 Conference on Robot Learning}, Proceedings of Machine Learning Research, 2021.

\bibitem[Mousavian et~al.(2019)Mousavian, Toshev, Fišer, Košecká, Wahid, and Davidson]{mousavian2019visualrep}
A.~Mousavian, A.~Toshev, M.~Fišer, J.~Košecká, A.~Wahid, and J.~Davidson.
\newblock Visual representations for semantic target driven navigation.
\newblock In \emph{2019 International Conference on Robotics and Automation (ICRA)}, 2019.

\bibitem[Maksymets et~al.(2021)Maksymets, Cartillier, Gokaslan, Wijmans, Galuba, Lee, and Batra]{maksymets2021thda}
O.~Maksymets, V.~Cartillier, A.~Gokaslan, E.~Wijmans, W.~Galuba, S.~Lee, and D.~Batra.
\newblock Thda: Treasure hunt data augmentation for semantic navigation.
\newblock In \emph{2021 IEEE/CVF International Conference on Computer Vision (ICCV)}, 2021.

\bibitem[Ye et~al.(2021)Ye, Batra, Das, and Wijmans]{ye2021auxiliary}
J.~Ye, D.~Batra, A.~Das, and E.~Wijmans.
\newblock Auxiliary tasks and exploration enable objectgoal navigation.
\newblock In \emph{2021 IEEE/CVF International Conference on Computer Vision (ICCV)}, 2021.

\bibitem[Majumdar et~al.(2022)Majumdar, Aggarwal, Devnani, Hoffman, and Batra]{majumdar2022zson}
A.~Majumdar, G.~Aggarwal, B.~Devnani, J.~Hoffman, and D.~Batra.
\newblock Zson: Zero-shot object-goal navigation using multimodal goal embeddings.
\newblock In \emph{Neural Information Processing Systems (NeurIPS)}, 2022.

\bibitem[Hu et~al.(2023)Hu, Xie, Jain, Francis, Patrikar, Keetha, Kim, Xie, Zhang, Zhao, Chong, Wang, Sycara, Johnson-Roberson, Batra, Wang, Scherer, Kira, Xia, and Bisk]{hu2023robotfmsurvey}
Y.~Hu, Q.~Xie, V.~Jain, J.~Francis, J.~Patrikar, N.~Keetha, S.~Kim, Y.~Xie, T.~Zhang, S.~Zhao, Y.~Q. Chong, C.~Wang, K.~Sycara, M.~Johnson-Roberson, D.~Batra, X.~Wang, S.~Scherer, Z.~Kira, F.~Xia, and Y.~Bisk.
\newblock Toward general-purpose robots via foundation models: A survey and meta-analysis, 2023.

\bibitem[Huang and Chang(2022)]{huang2022survey}
J.~Huang and K.~C.-C. Chang.
\newblock Towards reasoning in large language models: A survey.
\newblock \emph{arXiv preprint arXiv:2212.10403}, 2022.

\bibitem[Cai et~al.(2023)Cai, Huang, Cheng, Long, Gao, Sun, and Dong]{cai2023pixnav}
W.~Cai, S.~Huang, G.~Cheng, Y.~Long, P.~Gao, C.~Sun, and H.~Dong.
\newblock Bridging zero-shot object navigation and foundation models through pixel-guided navigation skill, 2023.

\bibitem[Honerkamp et~al.(2024)Honerkamp, Büchner, Despinoy, Welschehold, and Valada]{honerkamp2024momallm}
D.~Honerkamp, M.~Büchner, F.~Despinoy, T.~Welschehold, and A.~Valada.
\newblock Language-grounded dynamic scene graphs for interactive object search with mobile manipulation.
\newblock \emph{arXiv preprint arXiv:2403.08605}, 2024.

\bibitem[Amiri et~al.(2022)Amiri, Chandan, and Zhang]{amiri2022reasoningscenegraphs}
S.~Amiri, K.~Chandan, and S.~Zhang.
\newblock Reasoning with scene graphs for robot planning under partial observability.
\newblock \emph{IEEE Robotics and Automation Letters}, 2022.

\bibitem[Ravichandran et~al.(2022)Ravichandran, Peng, Hughes, Griffith, and Carlone]{ravichandran2022hierarchical}
Z.~Ravichandran, L.~Peng, N.~Hughes, J.~D. Griffith, and L.~Carlone.
\newblock Hierarchical representations and explicit memory: Learning effective navigation policies on 3d scene graphs using graph neural networks.
\newblock In \emph{2022 International Conference on Robotics and Automation (ICRA)}, 2022.

\bibitem[Ray et~al.(2024)Ray, Bradley, Carlone, and Roy]{ray2024tampscenegraphs}
A.~Ray, C.~Bradley, L.~Carlone, and N.~Roy.
\newblock Task and motion planning in hierarchical 3d scene graphs, 2024.

\bibitem[Strader et~al.(2024)Strader, Hughes, Chen, Speranzon, and Carlone]{strader2024ontology}
J.~Strader, N.~Hughes, W.~Chen, A.~Speranzon, and L.~Carlone.
\newblock Indoor and outdoor 3d scene graph generation via language-enabled spatial ontologies.
\newblock \emph{IEEE Robotics and Automation Letters}, 2024.

\bibitem[Werby et~al.(2024)Werby, Huang, Büchner, Valada, and Burgard]{werby2024hovsg}
A.~Werby, C.~Huang, M.~Büchner, A.~Valada, and W.~Burgard.
\newblock Hierarchical open-vocabulary 3d scene graphs for language-grounded robot navigation.
\newblock \emph{Robotics: Science and Systems}, 2024.

\bibitem[Mitra et~al.(2023)Mitra, Corro, Mahajan, Codas, Simoes, Agarwal, Chen, Razdaibiedina, Jones, Aggarwal, Palangi, Zheng, Rosset, Khanpour, and Awadallah]{mitra2023orca}
A.~Mitra, L.~D. Corro, S.~Mahajan, A.~Codas, C.~Simoes, S.~Agarwal, X.~Chen, A.~Razdaibiedina, E.~Jones, K.~Aggarwal, H.~Palangi, G.~Zheng, C.~Rosset, H.~Khanpour, and A.~Awadallah.
\newblock Orca 2: Teaching small language models how to reason, 2023.

\bibitem[Ren et~al.(2023)Ren, Dixit, Bodrova, Singh, Tu, Brown, Xu, Takayama, Xia, Varley, Xu, Sadigh, Zeng, and Majumdar]{ren2023knowno}
A.~Z. Ren, A.~Dixit, A.~Bodrova, S.~Singh, S.~Tu, N.~Brown, P.~Xu, L.~Takayama, F.~Xia, J.~Varley, Z.~Xu, D.~Sadigh, A.~Zeng, and A.~Majumdar.
\newblock Robots that ask for help: Uncertainty alignment for large language model planners.
\newblock In \emph{Proceedings of the Conference on Robot Learning (CoRL)}, 2023.

\bibitem[Ren et~al.(2024)Ren, Clark, Dixit, Itkina, Majumdar, and Sadigh]{ren2024explore}
A.~Z. Ren, J.~Clark, A.~Dixit, M.~Itkina, A.~Majumdar, and D.~Sadigh.
\newblock Explore until confident: Efficient exploration for embodied question answering, 2024.

\bibitem[Touvron et~al.(2023)Touvron, Martin, Stone, Albert, Almahairi, Babaei, Bashlykov, Batra, Bhargava, Bhosale, Bikel, Blecher, Ferrer, Chen, Cucurull, Esiobu, Fernandes, Fu, Fu, Fuller, Gao, Goswami, Goyal, Hartshorn, Hosseini, Hou, Inan, Kardas, Kerkez, Khabsa, Kloumann, Korenev, Koura, Lachaux, Lavril, Lee, Liskovich, Lu, Mao, Martinet, Mihaylov, Mishra, Molybog, Nie, Poulton, Reizenstein, Rungta, Saladi, Schelten, Silva, Smith, Subramanian, Tan, Tang, Taylor, Williams, Kuan, Xu, Yan, Zarov, Zhang, Fan, Kambadur, Narang, Rodriguez, Stojnic, Edunov, and Scialom]{touvron2023llama}
H.~Touvron, L.~Martin, K.~Stone, P.~Albert, A.~Almahairi, Y.~Babaei, N.~Bashlykov, S.~Batra, P.~Bhargava, S.~Bhosale, D.~Bikel, L.~Blecher, C.~C. Ferrer, M.~Chen, G.~Cucurull, D.~Esiobu, J.~Fernandes, J.~Fu, W.~Fu, B.~Fuller, C.~Gao, V.~Goswami, N.~Goyal, A.~Hartshorn, S.~Hosseini, R.~Hou, H.~Inan, M.~Kardas, V.~Kerkez, M.~Khabsa, I.~Kloumann, A.~Korenev, P.~S. Koura, M.-A. Lachaux, T.~Lavril, J.~Lee, D.~Liskovich, Y.~Lu, Y.~Mao, X.~Martinet, T.~Mihaylov, P.~Mishra, I.~Molybog, Y.~Nie, A.~Poulton, J.~Reizenstein, R.~Rungta, K.~Saladi, A.~Schelten, R.~Silva, E.~M. Smith, R.~Subramanian, X.~E. Tan, B.~Tang, R.~Taylor, A.~Williams, J.~X. Kuan, P.~Xu, Z.~Yan, I.~Zarov, Y.~Zhang, A.~Fan, M.~Kambadur, S.~Narang, A.~Rodriguez, R.~Stojnic, S.~Edunov, and T.~Scialom.
\newblock Llama 2: Open foundation and fine-tuned chat models, 2023.

\end{thebibliography}

\clearpage
\appendix
\renewcommand{\algorithmiccomment}[1]{\hspace{1em}\textcolor{lightgray}{$\triangleright$ #1}}
\renewcommand{\algorithmicdo}{}
\renewcommand{\algorithmicthen}{}
\renewcommand\algorithmicrequire{\textbf{Input:}}
\renewcommand\algorithmicensure{\textbf{Output:}}
\newcommand\tcb@cnt@specboxautorefname{\graphshort{} Spec}
\newcommand\tcb@cnt@llmpromptboxautorefname{LLM Prompt}
\newcommand\tcb@cnt@vqapromptboxautorefname{VQA Prompt}

\newcommand{\algorithmautorefname}{Algorithm}

\definecolor{forestgreen}{rgb}{0.13, 0.55, 0.13}

% Define a custom tcolorbox style for LLM prompts
\tcbset{
  specstyle/.style={
    % colback=lightgray,  % Background color
    colframe=darkgray, % Frame color
    coltitle=white, % Title text color
    fonttitle=\bfseries, % Title font
    boxrule=0.5mm, % Border thickness
    rounded corners,
    sharp corners=south, % Sharp bottom corners (optional)
    width=\linewidth, % Box width
    % arc=4mm, % Rounded corner radius
    % outer arc=2mm, % Outer rounded corner radius
    boxsep=5pt, % Box padding
    left=10pt, % Left padding
    right=10pt, % Right padding
    top=10pt, % Top padding
    bottom=10pt, % Bottom padding
    before=\bigskip, % Vertical space before the box
    after=\bigskip, % Vertical space after the box
    breakable,
  }
}

\newcommand\jsonkey{\color{purple}}
\newcommand\jsonvalue{\color{blue}}
\newcommand\jsonnumber{\color{brown}}

% switch used as state variable
\makeatletter
\newif\ifisvalue@json
\newif\ifisattr@json

\lstdefinelanguage{json}{
    tabsize             = 4,
    showstringspaces    = false,
    keywords            = {false,true},
    alsoletter          = 0123456789.,
    morestring          = [s]{"}{"},
    stringstyle         = \jsonkey\ifisattr@json\jsonnumber\fi\ifisvalue@json\jsonvalue\fi,
    MoreSelectCharTable = \lst@DefSaveDef{`[}\leftbracket@json{\enterItemMode@json},
    MoreSelectCharTable = \lst@DefSaveDef{`]}\rightbracket@json{\exitItemMode@json{\rightbracket@json}},
    MoreSelectCharTable = \lst@DefSaveDef{`\{}\leftbrace@json{\enterAttrMode@json},
    MoreSelectCharTable = \lst@DefSaveDef{`\}}\rightbrace@json{\exitAttrMode@json{\rightbrace@json}},
    basicstyle          = \ttfamily
}

\newcommand\enterItemMode@json{%
    \leftbracket@json%
    \ifnum\lst@mode=\lst@Pmode%
        \global\isvalue@jsontrue%
    \fi
}

\newcommand\exitItemMode@json[1]{#1\global\isvalue@jsonfalse}

\newcommand\enterAttrMode@json{%
    \leftbrace@json%
    \ifnum\lst@mode=\lst@Pmode%
        \global\isattr@jsontrue%
    \fi
}

\newcommand\exitAttrMode@json[1]{#1\global\isattr@jsonfalse}

\lst@AddToHook{Output}{%
    \ifisvalue@json%
        \ifnum\lst@mode=\lst@Pmode%
            \def\lst@thestyle{\jsonnumber}%
        \fi
    \fi
    %override by keyword style if a keyword is detected!
    \lsthk@DetectKeywords% 
}

\makeatother

\lstdefinestyle{Prompt}{
  basicstyle        = \small\ttfamily,
  showstringspaces  = false,
  breaklines        = true,
  breakindent       = 0pt,
  rulecolor         = \color{black},
  escapeinside      = {(*@}{@*)},
  moredelim         = [is][\color{purple}\bfseries]{@@}{@@},
  moredelim         = [is][\color{teal}\bfseries]{(@}{@)},
  moredelim         = [is][\bfseries]{**}{**},
  moredelim         = [is][\color{red}\itshape]{|}{|},
}

\newtcolorbox[auto counter]{specbox}[2][]{%
    title=\graphshort{} Spec~\thetcbcounter: #2, #1}

\newtcolorbox[auto counter]{llmpromptbox}[2][]{%
    title=LLM Prompt ~\thetcbcounter: #2, #1}

\newtcolorbox[auto counter]{vqapromptbox}[2][]{%
    title=VQA Prompt ~\thetcbcounter: #2, #1}

\section{System details}
\label{app:system_details}
\subsection{System hyperparameters}

In \autoref{tab:hyperparams}, we list the various modules in the system, describe the hyperparameters used in each and provide the values used in \navsys{} for our experiments. The \engineshort{}'s and Reasoner's hyperparameters are kept the same across both simulation and real-world tests, and across robot platforms in the real world. Some of the thresholds are sensor-dependent and can benefit from tuning for a specific sensor - \ie{} centroid nearness and minimum object size thresholds that are expressed in pixel-based units, and hence depend on image resolution or field-of-view. In practice, our centroid nearness threshold is defined to be large enough that sufficient object features can be accumulated when using both RGB sensors in simulation and on our robots in the real world.

\begin{table}
    \caption{\textbf{Hyperparameters used in the \navsys{} system.}}
    \begin{center}
        \begin{tabular}{@{\extracolsep{4pt}}lll@{}}
        \toprule
        \textbf{Subsystem} & \textbf{Hyperparameter} & \textbf{Value}  \\
        \midrule
        \multirow{4}{*}{\begin{tabular}{l}
            \engineshort{}
        \end{tabular}}& Centroid nearness \centroidthresh{} (object nearness heuristic) & 100 (pixels) \\
        & Bounding box overlap \bboxthresh{} (object nearness heuristic) & 0.1 (IoU)  \\
        & Min. object size threshold & 200 (pixels$^{2}$) \\
        & LLM temperature & 0.3 \\
        \midrule
        \multirow{1}{*}{\begin{tabular}{l}
            Reasoner
        \end{tabular}}& LLM temperature & 0.3 \\
        \midrule
        \multirow{5}{*}{\begin{tabular}{l}
            Controller\\
            (Simulation - FMM)
        \end{tabular}}& Occupancy map resolution & 0.05 (m) \\
        & Occupancy map size & 4.8 (m) \\
        & Min. obstacle height & 0.5 (m) \\
        & Turning action angle & 30 (deg) \\
        & Forward action distance & 0.25 (m) \\
        \midrule
        \multirow{3}{*}{\begin{tabular}{l}
             Controller\\
             (Real-world - GNM)
        \end{tabular}}& Stopping threshold & 2.25 \\
        & Max linear velocity & 0.5 (m/s) \\
        & Max angular velocity & 0.3 (rad/s) \\
        \bottomrule
        \end{tabular}
    \label{tab:hyperparams}
    \end{center}
\end{table}

\subsection{System runtime performance}
\begin{table}[tbp]
    \caption{\textbf{Statistics on foundation model queries made by \textit{\ourspr{}} system.} }
    \begin{center}
        \begin{tabular}{cccc}
        \toprule
        & & \multicolumn{2}{c}{Avg. no. of queries per episode} \\
        \cmidrule{3-4}
        \textbf{Model} & \textbf{Function} & \textbf{HM3D} & \textbf{Gibson}\\
        \midrule
        \multirow{1}{*}{VQA (BLIP2)} & \textit{Image parser} & 67.544 & 84.936 \\
        \midrule
        Object Detector (GroundingDINO) & \textit{Image parser} & 7.836 & 7.592 \\
        \midrule
        \multirow{3}{*}{LLM (OpenAI GPT-3.5)} & \textit{State estimator} & 58.718 & 45.804 \\
        & \textit{\graphshort{} updater}& 41.091 & 33.094 \\
        & \textit{Reasoner} & 50.739 & 45.455 \\
        \bottomrule                        
        \end{tabular}
    \label{tab:querytime}
    \end{center}
\end{table}

\begin{table}[tbp]
\caption{\textbf{\graphshort{} size statistics.} We quantify the number of nodes and edges in a typical \graphshort{} constructed in \hm{} scenes by \textit{\oursprgt{}}.}
\begin{center}
\begin{tabular}{@{\extracolsep{4pt}}cccccc@{}}
\toprule
\multicolumn{3}{c}{Nodes} & \multicolumn{3}{c}{Edges} \\
\cmidrule{1-3}  \cmidrule{4-6}
\textbf{Places} & \textbf{Structures} & \textbf{Objects} & \edgecontains{} & \edgeconnects{} & \edgenear{} \\
\midrule
3.030  & 8.476  & 81.253 & 84.283  & 18.915  & 18.729 \\
\bottomrule                        
\end{tabular}
\label{tab:num_nodesedges}
\end{center}
\end{table}

We provide additional analysis of \navsys{}'s runtime performance. As noted in \autoref{sec:conclusion}, \navsys{} can take 0.5-2 minutes to complete a full state estimation, \graphshort{} update and planning step. Most of the latency in this step is due to the need for multiple queries to the GPT-3.5 LLM. Once a plan is generated, the ViNT GNM can be run real-time ($\sim10$ Hz) to navigate toward the goal. We note that the other models used (BLIP-2 VQA model and GroundingDINO object detector) can both handle individual queries in $<0.5$ seconds on an AGX Orin. With batching of queries, these models can be used for inference at a close-to-real-time rate. To give a sense of the amount of computation required to perform ObjectNav, \autoref{tab:querytime} gives a breakdown of the mean number of queries to the various foundation models (excepting the GNM) per ObjectNav navigation episode. We address potential concerns about computational expense and discuss future directions in \autoref{app:add_discussion}.

We also provide details on the size of constructed \graphshort{}s in a typical \graphshort{} navigation episode in \autoref{tab:num_nodesedges}.

\section{\graphshort{} details}
\label{app:osg_details}

\subsection{\graphshort{} meta-structure details}

\textbf{Layer 1: Objects.} \textit{(Required).} Objects are leaf nodes in an \graphshort{}. Each object node \objnode{} represents a distinct object, and should at least contain the following 4 attributes: (i) an open-vocabulary text label for the object, \objnodelabel{} (\eg{} \textit{coffeetable, armchair}); (ii) an open-vocabulary text description of the object's appearance, \objnodetext{} (\eg{} \textit{white wooden} for a \textit{coffeetable} node); (iii) a unique node ID, \objnodeid{}; (iv) an image crop of the object, \objnodeimage{}. Object nodes are allowed to have outgoing \edgenear{} edges to other Object/Connector nodes.

\textbf{Layer 2: Connectors.} \textit{(Optional).} Connectors are also leaf nodes and share the same node attributes and allowed edges as Objects. In addition, they can be distinguished from Objects since they also capture spatial connectivity - \ie{} they are allowed to be connected to spatial region nodes via \edgeconnects{} edges. The Connectors layer is optional, since some environments - \eg{} open-plan offices - do not have Connectors connecting different spaces.

\textbf{Layer 3: Places.} \textit{(Required).} This layer contains Place nodes, \ie{} the finest-resolution spatial regions. To handle environments containing diverse kinds of spatial regions - \eg{} libraries consisting of both rooms (reading, seminar rooms) and aisles - we allow multiple \textit{classes} of Place nodes corresponding to the different kinds of regions to be specified. Each Place node \placenode{} contains the attributes: (i) its class as a text string drawn from the \graphshort{} specification, \placenodeclass{} (\eg{} rooms, aisles); (ii) an open-vocabulary text label describing the Place \placenodelabel{}, (\eg{} \textit{living room} for rooms, \textit{dairy aisle} for aisles); (iii) a unique node ID, \placenodeid{}. Valid outgoing edges for Place nodes are \edgecontains{} edges to Object nodes and \edgeconnects{} edges to Connector nodes.

\textbf{Layers 4-$N$: Region Abstractions.} \textit{(Optional).} Each node should contain the attributes: (i) an open-vocabulary text label describing the region \absinodelabel{}; (ii) a unique node ID, \absinodeid{}. Nodes in the $i$th layer may connect to nodes in the $(i-1)$th layer through \edgecontains{} edges. They may also connect to Connector nodes via \edgeconnects{} edges - \eg{} in \autoref{fig:splash} floors (Region Abstractions) may be linked through steps (Connector).

\subsection{\graphshort{} specification}
\begin{specbox}[specstyle, label=listing:abstract_spec]{Abstract example of an \graphshort{} specification)}
\begin{lstlisting}[language=json]
...
"AbstractionClass2": {
    "layer_type": "Region Abstraction",
    "layer_id": 5,
    "contains": ["AbstractionClass1"],
}
"AbstractionClass1": {
    "layer_type": "Region Abstraction",
    "layer_id": 4,
    "contains": ["PlaceClass1"],
    "connects to": ["ConnectorClass"]
},
"PlaceClass1": {
    "layer_type": "Place",
    "layer_id": 3,
    "contains": ["Object"],
    "connects to": ["PlaceClass1", "ConnectorClass"]
},
"ConnectorClass": {
    "layer_type": "Connector",
    "layer_id": 2,
    "is near": ["Object"],
    "connects to": ["PlaceClass", "AbstractionClass1"]
},
"Object": {
    "layer_id": 1
},
"State": ["PlaceClass1"]
\end{lstlisting}
\end{specbox}

We provide an example of an abstract \graphshort{} specification, formatted as a JSON string. This specification adheres to the \textit{meta-structure} described in \autoref{sec:approach_osg}. We provide examples in later sections for classes of real environments, \eg{} household scenes (\autoref{listing:sim_homes}), supermarkets (\autoref{listing:osg_spec_supermarket}) and several others. When fully instantiated, an \graphshort{} specification describes the structure and spatial concepts present in a specific class of environment.

\section{\enginelong{} details}
\label{app:mapper_details}

\subsection{Algorithm details}

\begin{algorithm}
\caption{\textit{\engineshort{}}}
\label{alg:mapper}
\begin{algorithmic}[1]
    \REQUIRE Image $I_t$, \graphshort{} specification $\mathcal{S}$, previous \graphshort{} $\mathcal{G}$, previous robot state (Place) $P_{t-1}$
    \ENSURE Updated \graphshort{} $\mathcal{G}^{'}$, updated current state (Place) $P_t$
    \STATE $P, O, C \leftarrow$ \textsc{ImageParser}($I_t, \mathcal{S}$) \COMMENT{Identify Objects, Connectors and Places from RGB observations. See \autoref{alg:mapper_image_parser}}
    \STATE $\hat{P}_t\leftarrow$ \textsc{StateEstimator}($\mathcal{S}, \mathcal{G}, P_{t-1}, P, O, C$) \COMMENT{See \autoref{alg:mapper_state_est}}
    \STATE $\mathcal{G}^{'}, P_t\leftarrow$ \textsc{OSGUpdater}($\mathcal{S}, \mathcal{G}, \hat{P}_t, P_{t-1}, P, O, C$) \COMMENT{See \autoref{alg:mapper_osg_updater}}
    \RETURN $\mathcal{G}^{'}, P_t$
\end{algorithmic}
\end{algorithm}

\begin{algorithm}
\caption{\textit{Image Parser}}
\label{alg:mapper_image_parser}
\begin{algorithmic}[1]
    \REQUIRE RGB image $I_t$, \graphshort{} specification $\mathcal{S}$
    \ENSURE Extracted Place $P$, Objects $O$, and Connectors $C$ from observations
    \STATE $P \leftarrow$ \textcolor{blue}{\textsc{LabelPlace\_VQA}}($I_t, \mathcal{S}$) \COMMENT{See \autoref{listing:place_class_struct}}
    \STATE $D \leftarrow$ \textcolor{blue}{\textsc{DetectObjectsConnectors\_ObjDet}}($I$) \COMMENT{GroundingDINO open-set detection}
    \STATE $O, C \leftarrow$ \textcolor{blue}{\textsc{ClassifyObjectsConnectors\_LLM}}($D, \mathcal{S}$) \COMMENT{See \autoref{listing:classify_elements_struct}}
    \FOR{$o$ \textbf{in} $O \cup C$:}
        \STATE $o.attr \leftarrow$ \textcolor{blue}{\textsc{LabelWithTextualAttribs\_VQA}}($o$) \COMMENT{See \autoref{listing:obj_description_struct}}
    \ENDFOR
    \RETURN $P, O, C$
\end{algorithmic}
\end{algorithm}

\begin{algorithm}
\caption{\textit{State Estimator}}
\label{alg:mapper_state_est}
\begin{algorithmic}[1]
\REQUIRE \graphshort{} specification $\mathcal{S}$, \graphshort{} $\mathcal{G}$, previous robot state $P_{t-1}$, observed Place $P$, Objects $O$, Connectors $C$
\ENSURE Estimated robot state (Place node) $\hat{P}_t$

\STATE place\_nodes $\leftarrow$ \textsc{GetLayer}($\mathcal{G}, 3$) \COMMENT{Get all nodes from Layer 3: Places}
\STATE similar\_places $\leftarrow$ \textcolor{blue}{\textsc{GetSimilarPlaces\_LLM}}($\mathcal{S}, P_t$, place\_nodes) \COMMENT{Identify Place nodes with semantically similar names (\eg{} all``living rooms'') See \autoref{listing:place_name_struct}}
\STATE sorted\_places $\leftarrow$ \textsc{SortByDistance}(similar\_places, $P_{t-1}$) \COMMENT{Order similar places by increasing distance from last robot state}
\FOR{place \textbf{in} sorted\_places:}
    \STATE stored\_feats $\leftarrow$ \textsc{GetObjFeats}($\mathcal{G}$, place) \COMMENT{Object features of \graphshort{} place node}
    \IF{\textcolor{blue}{\textsc{PairwisePlaceMatch\_LLM}}($\mathcal{S}, O \cup C$, stored\_feats): \COMMENT{See \autoref{listing:place_recognition_struct}}}
        \RETURN place
    \ENDIF
\ENDFOR
\RETURN \texttt{None}
\end{algorithmic}
\end{algorithm}

\begin{algorithm}
\caption{\textit{OSG Updater}}
\label{alg:mapper_osg_updater}
\begin{algorithmic}[1]
\REQUIRE \graphshort{} specification $\mathcal{S}$, \graphshort{} $\mathcal{G}$, estimated state $\hat{P}_{t}$, previous state $P_{t-1}$, observed Places $P$, Objects $O$, Connectors $C$
\ENSURE Updated \graphshort{} $\mathcal{G}^{'}$, updated current state $P_t$
\IF{$\hat{P}_t$ \textbf{is} \texttt{None}:}
    \STATE \textcolor{forestgreen}{// Robot is in a new Place. Add new nodes/edges to Layers $1-3$.}
    \STATE $P_t \leftarrow \textsc{AddPlaceNode}(\mathcal{G}, \hat{P}_t)$
    \STATE \textsc{AddLeafNodes}($\mathcal{S}, \mathcal{G}, O\cup C$) \COMMENT{Add observed Objects/Connectors, and contains/connects edges following structure specified in $\mathcal{S}$}
    \STATE \textsc{AddIsNearEdges}($\hat{P}_t, O\cup C$) \COMMENT{Add \edgenear{} edges among leaf nodes}

    \STATE \textcolor{forestgreen}{// Update abstract regions, from Layers $4-N$.}
    \STATE curr\_region, prev\_region $\leftarrow\hat{P}_t, P_{t-1}$
    \STATE $N\leftarrow$ \textsc{GetNumLayers}($\mathcal{S}$)
    \FOR{$i$ in \texttt{range}(4, $N+1$): \COMMENT{Iterates from Layer $4$ to Layer $N$}}
        \STATE layer $\leftarrow$ \textsc{GetLayer}($\mathcal{G}, i$) \COMMENT{Get all nodes of layer $i$}
        \STATE parent\_region $\leftarrow$ \textcolor{blue}{\textsc{InferAbstractRegion\_LLM}}($\mathcal{S}, i$, curr\_region, prev\_region, layer) \COMMENT{Based on current region's (layer $i-1$) description and agent's location history, suggest parent region in layer $i$. See \autoref{listing:region_abs_struct}}
        \IF{parent\_region \textbf{in} layer\_nodes:}
            \STATE \textsc{AddContainsEdge}($\mathcal{S}, \mathcal{G}$, parent\_region, curr\_region)
            \STATE \textbf{break}
        \ELSE
            \STATE prev\_parent\_region $\leftarrow$ \textsc{GetParentNode}($\mathcal{G}$, prev\_region)
            \STATE \textsc{AddRegionNode}($\mathcal{S}, \mathcal{G}$, parent\_region) \COMMENT{Add new node to layer $i$}
            \STATE \textsc{AddEdges}($\mathcal{S}, \mathcal{G}$, parent\_region, curr\_region, prev\_parent\_region) \COMMENT{Add edges between new node and its neighbour/child nodes, following structure in $\mathcal{S}$}
        \ENDIF
    \ENDFOR
    \STATE curr\_region $\leftarrow$ parent\_region
    \STATE prev\_region $\leftarrow$ prev\_parent\_region
\ELSE
    \STATE \textcolor{forestgreen}{// Robot in existing Place node. We add/update Layers $1-2$ nodes in current Place node.}
    \STATE $P_t\leftarrow \hat{P}_t$
    \STATE $\mathcal{V}_{leaf} \leftarrow$ \textsc{GetLeafNodes}($\mathcal{G}$, $\hat{P}_t$) \COMMENT{Get objects, connectors contained in current Place}
    \STATE $\mathcal{F}_{leaf} \leftarrow$ \texttt{map}(\textsc{GetObjFeats}, $\mathcal{V}_{leaf}$) \COMMENT{Get object features for each leaf node in $\mathcal{V}_{leaf}$}
    \FOR{$o$ \textbf{in} $O \cup C$:}
        \STATE $f_o \leftarrow$ \textsc{GetObjFeats}($o$)
        \STATE matched\_obj $\leftarrow$ \textcolor{blue}{\textsc{AssociateObject\_LLM}}($o, f_o, \mathcal{V}_{leaf}, \mathcal{F}_{leaf}$) \COMMENT{See \autoref{listing:data_assoc_ex}}
        \IF{matched\_obj \textbf{is} \texttt{None}}
            \STATE \textsc{AddLeafNodes}($\mathcal{S}, \mathcal{G}, \{o\}$) \COMMENT{Add $o$ and contains/connects edges following $\mathcal{S}$}
        \ELSE
            \STATE \textsc{UpdateLeafNode}($\mathcal{G}, o$)
            \STATE \textsc{UpdateEdges}($\mathcal{G}, o$)
        \ENDIF
    \ENDFOR
\ENDIF
\RETURN $\mathcal{G}, P_t$
\end{algorithmic}
\end{algorithm}

We provide additional details on the \engineshort{} mapper algorithm from \autoref{sec:approach_mapper}. \autoref{alg:mapper} provides an overview of the abstract algorithm sketch. The sketch takes an \graphshort{} specification as an argument, which is used to instantiate it into an executable routine for a specific class of environments. The specification contains open-vocabulary spatial concepts relevant to that environment, which are used to fill in the templated prompts to LLMs/VLMs during map-building. It also contains specific structural information (\ie{} allowable edges) used during \graphshort{} construction.

We label the functions used to prompt foundation models in \textcolor{blue}{\textsc{blue}}, and provide examples of the prompts in the following subsections. If a prompt is templated, we show the templated version with placeholder values highlighted in \textcolor{red}{red} followed by an instantiated example assuming an \graphshort{} specification for a common household environment, such as can be found in \autoref{listing:sim_homes}.

\subsection{Image Parser prompts}
\begin{vqapromptbox}[specstyle, label=listing:place_class_struct]{Prompt for Place Labelling}
\begin{lstlisting}[style=Prompt]
What |<PlaceClass>| are we in?
\end{lstlisting}
\end{vqapromptbox}

\begin{vqapromptbox}[specstyle, label=listing:obj_description_struct]{Prompt for Object/Connector Appearance Description}
\begin{lstlisting}[style=Prompt]
What color is the |<object>|/|<connector>| in the image?

What material is the |<object>|/|<connector>| made of?
\end{lstlisting}
\end{vqapromptbox}

\begin{llmpromptbox}[specstyle, label=listing:classify_elements_struct]{Templated prompt for Scene Element Classification}
\begin{lstlisting}[style=Prompt]
@@Few-shot prompts@@
(*@\textbf{<Few-shot prompts follow format of query shown below. Examples of few-shot prompts given in \autoref{listing:classify_elements_struct}>}@*)

@@Query@@
We observe the following: ["livingroom_0", "window_13", ...]. Please eliminate redundant strings in the element from the list and classify them into |<PlaceClass>|, |<ConnectorClass>|, Object classes.
**Answer**: 
|<PlaceClass>|: ... 
|<ConnectorClass>|: ...
**object**: ...
\end{lstlisting}
\end{llmpromptbox}

\begin{llmpromptbox}[specstyle, label=listing:classify_elements_ex]{Example interaction with LLM for Scene Element Classification}
\begin{lstlisting}[style=Prompt]
@@Few-shot prompts@@
There is a list: ["bathroom_0", "bathroom mirror_1","bathroom sink_2","toilet_3", "bathroom bathtub_4", "lamp_5", "ceiling_10"]. Please eliminate redundant strings in the element from the list and classify them into "room," "entrance," and "object" classes. Ignore floor, ceiling and wall.

**Answer**: 
room: bathroom_0
entrance: none
object: mirror_1, sink_2, toilet_3, bathtub_4, lamp_5

...

@@Example query@@
We observe the following: ["livingroom_0", "window_13","door_2", "doorway_3", "table_4","chair_5","livingroom sofa_6", "floor_7", "wall_8", "doorway_9", "stairs_10", "tv_16", "stool_17", "couch_18", "remote_19"]. Please eliminate redundant strings in the element from the list and classify them into room, doorway, object classes.

(@Example LLM reply@)
**Answer**:
room: livingroom_0
entrance: door_2, doorway_3, doorway_9
stair: stairs_10
object: table_4, chair_5, sofa_6, window_13, tv_16, stool_17, couch_18, remote_19
\end{lstlisting}
\end{llmpromptbox}

\subsection{State Estimator prompts}

\begin{llmpromptbox}[specstyle, label=listing:place_name_struct]{Templated prompt for Place Label Similarity}
\begin{lstlisting}[style=Prompt]
**Question**: In the scene, there are ["kitchen_1", "livingroom_1", "livingroom_2", "bedroom_1", "familyroom_2"]. Now, I need to find a "livingroom". Please give me all the |<PlaceClass>| with similar semantic meaning from the list. Please directly return the answer in a Python list. Follow the format: Answer: <your answer>." 
\end{lstlisting}
\end{llmpromptbox}

\begin{llmpromptbox}[specstyle, label=listing:place_name_ex]{Example interaction with LLM for Place Label Similarity}
\begin{lstlisting}[style=Prompt]
@@Example query@@
**Question**: In the scene, there are ["kitchen_1", "livingroom_1", "livingroom_2", "bedroom_1", "familyroom_2"]. Now, I need to find a "livingroom". Please give me all the rooms with similar semantic meaning from the list. Please directly return the answer in a Python list. Follow the format: Answer: <your answer>." 

(@Example LLM reply@)
Answer: livingroom_1, livingroom_2, familyroom_2.
\end{lstlisting}
\end{llmpromptbox}

\begin{llmpromptbox}[specstyle, label=listing:place_recognition_struct]{Templated prompt for Pairwise Place Matching}
\begin{lstlisting}[style=Prompt]
@@Context@@
You are a robot exploring an environment for the first time. You will be given an object to look for and should provide guidance on where to explore based on a series of observations. Observations will be given as descriptions of objects seen from four cameras in four directions. Your job is to estimate the robot's state. You will be given two descriptions, and you need to decide whether these two descriptions describe the same |<PlaceClass>|. For example, if we have visited the |<PlaceClass>| before and got one description, when we visit a similar |<PlaceClass>| and get another description, it is your job to determine whether the two descriptions represent the same |<PlaceClass>|. You should understand that descriptions may contain errors and noise due to sensor noise and partial observability. Always provide reasoning along with a deterministic answer. If there are no suitable answers, leave the space after 'Answer: None.' Always include: Reasoning: <your reasoning> Answer: <your answer>.

@@Few-shot prompts@@
(*@\textbf{<Few-shot prompts follow format of query shown below. Examples of few-shot prompts given in \autoref{listing:place_recognition_ex}>}@*)

@@Query@@
**Description1**: I can see a brown wood headboard, ...
**Description2**: I can see a white wooden door, ...
**Question**: These are depictions of what I observe from two different vantage points. Please assess the shared objects and spatial relationship in the descriptions to determine whether these two positions are indeed in the same place. Provide a response of True or False, along with supporting reasons. In each direction, focus on only two to three large objects for reasoning.
**Answer**: _
**Reasoning**: _
\end{lstlisting}
\end{llmpromptbox}

\begin{llmpromptbox}[specstyle, label=listing:place_recognition_ex]{Example interaction with LLM for Pairwise Place Matching}
\begin{lstlisting}[style=Prompt]
@@Context@@
You are a robot exploring an environment for the first time...

@@Few-shot prompts@@
**Description1**: On the left, I can see a brown wood headboard, white paper pillow. On the right, I can see a black metal television, gray plastic laundry basket, white wood closet dresser, brown wood drawer. In fron of me, I can see a white wood bed, white wood window, brown metal lamp, brown wood dresser, brown wood dresser nightstand, black silk curtain, white plastic curtain, white metal wall lamp, brown wood drawer. Behind me, I can see a brown wood cabinet. 
**Description2**: On the left, I can see a white wood door. On the right, I can see a white wood bed, white glass lamp, white glass window, white plastic curtain, brown wood dresser nightstand, white glass window, white wood nightstand, blue fabric curtain, white cotton pillow, white metal ceiling fan, silver metal wall lamp. In front of me, I can see a white glass lamp, brown wood headboard, white cotton pillow, brown wood dresser, white wood bed, white cotton pillow, red metal wall lamp door, brown wood drawer. Behind me, I can see a brown wood bureau, black glass television, brown wood stool, brown wood drawer, brown wood drawer, brown wood drawer.
**Question**: These are depictions of what I observe from two different vantage points. Please assess the shared objects and spatial relationship in the descriptions to determine whether these two positions are indeed in the same place. Provide a response of True or False, along with supporting reasons. In each direction, focus on only two to three large objects for reasoning.
**Answer**: True
**Reasoning**: To simplify the description, given the abundance of objects, we initially focus on common perceptions of object sizes. Our attention is directed toward larger objects, as these are less prone to detection errors.
Description 1: On the left, there is brown wood headboard, white paper pillow. On the right, black metal television, brown wood dresser, and white wood closet dresser are of relatively large size. In front of me, there is a white wood bed and a brown wood dresser. Behind me, I can see a brown wood cabinet and black silk curtain.
Description 2: On the left, there is white wood door. On the right, the white wood bed, brown wood dresser, nightstand, and ceiling fan are of relatively large size. In front of me, there is a white wood bed, black glass television a brown wood dresser. Behind me, I can see a brown wood cabinet.
Shared Large Objects: the two descriptions exhibit significant commonalities, prominently featuring large and easily observable items such as a brown wood dresser, a brown wood drawer, a white wood bed, a metal wall lamp, a television, and a brown wood headboard. Spatial Relationship: the spatial relationships within both descriptions remain consistent, with the dresser and wall lamp positioned near the bed in each scenario. Despite minor variations in the color or material of smaller objects like stools or curtains, these discrepancies appear more likely to stem from observational nuances rather than indicating distinct rooms. 

...

@@Example query@@
**Description1**: On the left, there is a silver metal faucet, white glass mirror, black metal wall lamp, white stainless steel sink, white formica countertop, blue plastic accessory, white wood white door, silver metal faucet, white metal wall lamp, brown wood cabinet, black wood white door doorway, brown wood cabinet, white wood mirror wall. On the right, there is a brown tile floor, white drywall ceiling. In front of me, there is a silver metal faucet, silver glass mirror, white soap soap, white porcelain sink, brown wood cabinet, white porcelain sink, silver metal wall lamp, silver metal wall lamp. At the rear, there is an orange glass lamp, brown wood floor, white porcelain tub, white cotton bed.
**Description2**:  On the left, there is a white glass mirror, silver metal faucet, white porcelain sink, white wood bathroom sink, white wood bed, brown metal wall lamp. On the right, there are no specified items. In front of me, there is a silver metal faucet, white porcelain sink, white plastic toiletry, silver glass mirror, blue plastic soap toiletry, white white bathroom sink countertop, blue metal wall lamp, black wood bathroom cabinet, black metal wall lamp. Behind me, there is a white porcelain tub, black cloth curtain, white cotton bed, brown wood bed, black cloth curtain, white porcelain bath.
**Question**: These are depictions of what I observe from two different vantage points. Please assess the shared objects and spatial relationship in the descriptions to determine whether these two positions are indeed in the same place. Provide a response of True or False, along with supporting reasons. In each direction, focus on only two to three large objects for reasoning.

(@Example LLM reply@)
**Answer**: True
**Reasoning**: To simplify the description, given the abundance of objects, we focus initially on common perceptions of larger objects, as they are less prone to detection errors.
Description 1: On the left, there is brown wood cabinet, silver metal faucet, white wood bathroom sink, and white wood bed. On the right, there is nothing. In front of me, there is a silver metal faucet, silver glass mirror,white porcelain sink, brown wood cabinet, black mental wall lamp.  Behind me, there is an orange glass lamp, white porcelain tub, white cotton bed. 
Description 2: On the left, there is white porcelain sink and white wood bed. On the right, there are nothing. In front of me, there is a silver mental faucet and white plastic toiletry. Behind me, I can see white porcelain tub, black cloth curtain, white cotton bed.
Shared Large Objects: Both descriptions mention silver metal faucets, white beds, and sink, contributing to a consistent thematic presence. Spatial Relationship: The placement of key items like silver metal faucets and brown wood cabinets described in similar configurations create a sense of coherence. Potential for Small Errors: Description 2 lacks specified items, such as accessory, door and countertop. Besides, there are small errors in the material of bed. Errors in material or color, as well as the potential omission of small-size objects, can indeed be attributed to variations in observation. Despite this, the two descriptions are depicting the same room, with the potential for small errors accounting for minor variations.
\end{lstlisting}
\end{llmpromptbox}

\subsection{\graphshort{} Updater prompts}

\begin{llmpromptbox}[specstyle, label=listing:data_assoc_ex]{Example interaction with LLM for Object Data Association}
\begin{lstlisting}[style=Prompt]
@@Few-shot prompts@@
**Object to be matched**: We want to find a doorframe that is near a tv, a chair and a stool.
**List of observed objects**: Now we have seen the following object: doorframe_2 that is near chair and sofa. doorframe_3 that is near a tv and a chair. wooden door_2 that is near table, sink and lamp.
**Question**: Please select one object that is most likely to be the object I want to find. Please only select one object in the list and use this element name in answer. Use the exact name in the given sentences. Always follow the format: Answer: <your answer>."
**Answer**: doorframe_3
**Reasoning**: Among the given objects, "doorframe_3" is mentioned to be near a TV and a chair, most likely meeting the specified criteria of being near a TV, chair, and stool.

...

@@Example query@@
**Object to be matched**: We want to find a door that is near a dining table, window glass and a table cloth.
**List of observed objects**: Now we have seen the following object: door_2 that is near table and window glass. doorframe_4 that is near a table cloth and a glass. door_4 that is near photo, sofa and windows.
**Question**: Please select one object that is most likely to be the object I want to find. Please only select one object in the list and use this element name in answer. Use the exact name in the given sentences. Always follow the format: Answer: <your answer>."

(@Example LLM reply@)
**Answer**: door_2
**Reasoning**: Among the given objects, 'door_2' is mentioned to be near a table and window glass. Although 'doorframe_4' is also near a tablecloth and a glass, it seems to meet the criteria. However, since the target goal is a door and not a doorframe, 'door_2' is more suitable.
\end{lstlisting}
\end{llmpromptbox}

\begin{llmpromptbox}[specstyle, label=listing:region_abs_struct]{Templated prompt for Inferring Region Abstractions}
\begin{lstlisting}[style=Prompt]
Previously, we were in |<AbstractionClass>| containing the |<PlaceClass>|, and we move towards |<Subgoal>|. Now we arrive in a |<PlaceClass>|. Do you think the current state |<PlaceClass>| belongs to [list of existing |<AbstractionClass>|s]?

If it belongs to any other existing region abstraction, return the region abstraction name; otherwise propose the name for this new |<AbstractionClass>| formatted as "<name> |<AbstractionClass>| (New)". Ensure that your response follows the format:
Reasoning: <your reasoning>. 
Answer: <your answer>
\end{lstlisting}
\end{llmpromptbox}

\begin{llmpromptbox}[specstyle, label=listing:region_abs_ex]{Example interaction with LLM for Inferring Region Abstractions}
\begin{lstlisting}[style=Prompt]
@@Example query@@
Previously, we were in the Cooked Food Section containing the Deli Aisle, and we move towards Freezer. Now we arrive in a Dairy Aisle. Do you think the current state Dairy Aisle belongs to any of the existing region abstractions: ["Cooked Food Section", "Clothing Section", "Dry Goods Section"]? 

If it belongs to any other existing region abstraction, return the region abstraction name; otherwise propose the name for this new section formatted as "<name> Section (New)". Ensure that your response follows the format: 
Reasoning: <your reasoning>. 
Answer: <your answer>

(@Example LLM reply@)
Reasoning: The Dairy Aisle typically contains products such as milk, cheese, yogurt, and butter, which are perishable food items. Given its nature as a section primarily housing food products, it doesn't fit within the existing region abstractions provided ("Cooked Food Section", "Clothing Section", "Dry Goods Section").

Answer: Perishable Food Section (New)
\end{lstlisting}
\end{llmpromptbox}

\section{Reasoner details}
\label{app:reasoner_details}

\subsection{Algorithm details}
% \begin{algorithm}
% \caption{Calculate $y = x^n$}
% \label{alg1}
% \begin{multicols}{2}
% \begin{algorithmic}[1]
%   \REQUIRE $n \geq 0 \vee x \neq 0$
%   \ENSURE $y = x^n$
%   \STATE $y \Leftarrow 1$
%   \IF{$n < 0$}
%   \STATE $X \Leftarrow 1 / x$
%   \STATE $N \Leftarrow -n$
%   \ELSE
%   \STATE $X \Leftarrow x$
%   \STATE $N \Leftarrow n$
%   \ENDIF
%   \WHILE{$N \neq 0$}
%   \IF{$N$ is even}
%   \STATE $X \Leftarrow X \times X$
%   \STATE $N \Leftarrow N / 2$
%   \ELSE[$N$ is odd]
%   \STATE $y \Leftarrow y \times X$
%   \STATE $N \Leftarrow N - 1$
%   \ENDIF
%   \ENDWHILE
% \end{algorithmic}
% \end{multicols}
% \end{algorithm}

\begin{algorithm}
\caption{Outline of \textit{Reasoner} algorithm}
\label{alg:reasoner}
\begin{algorithmic}[1]
    \REQUIRE \graphshort{} specification $\mathcal{S}$, \graphshort{} instance $\mathcal{G}$
    \ENSURE Execution of subgoals and pathfinding
    
    \STATE likely\_region $\leftarrow$ \texttt{None}
    \FOR{$i$ \textbf{in} \texttt{range}(N, 2): \COMMENT{Iterates from Layer $N$ to Layer 3 (Places)}}
        \IF{$i == N$}
            \STATE subset $\leftarrow$ \textsc{GetLayer}($\mathcal{G}, i$)
        \ELSE
            \STATE subset $\leftarrow$ \textsc{GetChildNodes}($\mathcal{G}$, likely\_region)
        \ENDIF
        \STATE layer\_class $\leftarrow$ \textsc{GetAbstractionClass}($\mathcal{S}, i$) \COMMENT{Get spatial concept of layer $i$}
        \STATE likely\_region $\leftarrow$ \textcolor{blue}{\textsc{RegionProposer\_LLM}}(subset, layer\_class) \COMMENT{See \autoref{listing:region_prompt_struct}}
    \ENDFOR
    
    \STATE path $\leftarrow$ \textsc{Pathfinder}($s$, likely\_region) \COMMENT{Find path (sequence of Places) over \graphshort{}}
    \WHILE{path \textbf{is not} \texttt{empty}:}
        \STATE next\_waypoint $\leftarrow$ path.\texttt{pop}()
        \IF{next\_waypoint \textbf{is a} Connector}
            \STATE \textsc{CommandControllerSubgoal}(next\_waypoint)
        \ELSIF{next\_waypoint \textbf{is} \texttt{Region}}
            \STATE objs\_in\_current\_node $\leftarrow$ \textsc{GetObjectsInNode}($\mathcal{G}, s$)
            \STATE subgoal $\leftarrow$ \textcolor{blue}{\textsc{GoalProposer\_LLM}}(objs\_in\_current\_node)  \COMMENT{See \autoref{listing:object_prompt_struct}}
        \ENDIF
        
        \STATE \textsc{WaitUntilSubgoalReached}()
        \STATE $\mathcal{G}, s \leftarrow$ \textsc{GetUpdatedOSGAndState}()
    \ENDWHILE
\end{algorithmic}
\end{algorithm}

We provide additional details on the Reasoner algorithm from \autoref{sec:approach_reasoner}. \autoref{alg:reasoner} provides an overview of the abstract algorithm sketch. The sketch takes an \graphshort{} specification as an argument, which is used to instantiate it into an executable routine for a specific class of environments. The specification contains open-vocabulary spatial concepts relevant to that environment, which are used to fill in the templated prompts to LLMs, and provide specific structural information (\ie{} hierarchy) to guide the order of LLM queries.

Intuitively, we prompt the LLM with the description of our object search task and the \graphshort{} to encourage it to form a belief of the target's location, then query this belief in a top-down manner. Region Proposer hierarchically queries the LLM to identify promising regions to search at each layer of the \graphshort{}. We also use Goal Proposer to query its belief and identify promising objects to search nearby.

We label the functions used to prompt foundation models in \textcolor{blue}{\textsc{blue}}, and provide examples of the prompts in the following subsections. If a prompt is templated, we show the templated version with placeholder values highlighted in \textcolor{red}{red} followed by an instantiated example assuming an \graphshort{} specification for a common household environment, such as can be found in \autoref{listing:sim_homes}.

\subsection{Region Proposer prompts}

\begin{llmpromptbox}[specstyle, label=listing:region_prompt_struct]{Templated prompt for Region Proposal}
\begin{lstlisting}[style=Prompt]
@@Few-shot prompts@@
(*@\textbf{<Few-shot prompts follow format of query shown below. Examples of few-shot prompts given in \autoref{listing:region_prompt_ex}>}@*)

@@Query@@
(*@\textbf{\graphlong{}}@*): You see the partial layout of the environment: {"room": {"livingroom_1": ...
**Goal**: Your goal is to find a |<Goal>| object.
**Question**: If any of the |<LayerType>| in the layout are likely to contain the target object, specify the most probable |<LayerType>| name. If all the |<LayerType>| are not likely to contain the target object, provide the |<ConnectorClass>| (that could connect <LayerType>) you would select for exploring a new |<LayerType>| where the target object might be found.
**Answer**: _
**Reasoning**: _
\end{lstlisting}
\end{llmpromptbox}

\begin{llmpromptbox}[specstyle, label=listing:region_prompt_ex]{Example interaction with LLM for Region Proposal}
\begin{lstlisting}[style=Prompt]
@@Few-shot prompts@@
(*@\textbf{\graphlong{}}@*): 
You see the partial layout of the apartment: {"room": {"kitchen_1": {"connects to": ["stair_1"]}, "livingroom": {"connects to": ["stair_1"]}, "entrance": {"stair_1": {"is near": []}}}}
**Goal**: Your goal is to find a sink.
**Question**: If any of the rooms in the layout are likely to contain the target object, specify the most probable room name. If all the room are not likely contain the target object, provide the door you would select for exploring a new room where the target object might be found.
**Answer**: kitchen_1
**Reasoning**: There are kitchen and livingroom in the layout. Among all the rooms, kitchen is usually likely to contain sink. Since we haven't explored the kitchen yet, it is possible that the sink is in the kitchen. Therefore, I will explore kitchen.

(*@\textbf{\graphlong{}}@*): 
You see the partial layout of the apartment: {"room": {"livingroom_1", "connects to": ["door_1", "door_2"]}, "diningroom_1": {,"connects to": ["door_1"]}}, "entrance": {"door_1": {"is near": ["towel_1"], "connects to": ["livingroom_1", "diningroom_1"]}, "door_2": {"is near": [], "connects to": ["livingroom_1"]}}}
**Goal**: Your goal is to find a sink.
**Question**: If any of the rooms in the layout are likely to contain the target object, specify the most probable room name. If all the room are not likely contain the target object, provide the door you would select for exploring a new room where the target object might be found.
**Answer**: door_1
**Reasoning**: There is only livingroom in the layout. livingroom is not likely to contain sink, so I will not explore the current room. Among all the doors, door1 is near to towel. A towel is usually more likely to near the bathroom or kitchen, so it is likely that if you explore door1 you will find a bathroom or kitchen and thus find a sink.

...

@@Example query@@
(*@\textbf{\graphlong{}}@*): 
You see the partial layout of the environment: {"room": {"livingroom_1": {"connects to": ["doorway_1", "door_2"]}, "entrance": {"doorway_1": {"is near": ["table_1"]}, "door_2": {"is near": ["clock_1"], "connects to": ["livingroom_1" ]}}} 
**Goal**: Your goal is to find a oven.
**Question**: If any of the rooms in the layout are likely to contain the target object, specify the most probable room name. If all the room are not likely contain the target object, provide the door you would select for exploring a new room where the target object might be found.

(@Example LLM reply@)
**Answer**: doorway_1
**Reasoning**: There are only livingroom in the layout. Among all the rooms, livingroom is usually unlikely to contain an oven, making it less likely for me to find an oven in the current room. Instead, I plan to explore other rooms connected to the current living room via entrances. Evaluating the entrances, doorway1 stands out as it is close to a table. Tables are commonly found in kitchens, which often contain ovens. Therefore, I have decided to explore through doorway_1.
\end{lstlisting}
\end{llmpromptbox}

\subsection{Goal Proposer prompts}

\begin{llmpromptbox}[specstyle, label=listing:object_prompt_struct]{Templated prompt for Goal Proposal}
\begin{lstlisting}[style=Prompt]
@@Few-shot prompts@@
(*@\textbf{<Few-shot prompts follow format of query shown below. Examples of few-shot prompts given in \autoref{listing:object_prompt_ex}>}@*)

@@Query@@
**Objects in current Place node**: There is a list: ["mirror_2", "lamp_1", ...]. 
**Question**: Please select one object that is most likely located near a |<Goal>|. Always follow the format: Answer: <your answer>.
**Answer**: _
**Reasoning**: _
\end{lstlisting}
\end{llmpromptbox}

\begin{llmpromptbox}[specstyle, label=listing:object_prompt_ex]{Example interaction with LLM for Goal Proposal}
\begin{lstlisting}[style=Prompt]
@@Few-shot prompts@@
**Objects in current Place node**:
There is a list: ["chair_4", "sofa_2", "bed_9", "dresser_1","ceiling_6","closet_5", "window_7", "wall_10"]. 
**Question**: Please select one object that is most likely located near a table. Always follow the format: Answer: <your answer>. Reasoning: <your Reasoning>.
**Answer**: chair_4
**Reasoning**: Among the given options, the object most likely located near a table is a "chair." Chairs are commonly placed around tables for seating during various activities such as dining, working, or socializing.
...

@@Query@@
**Objects in current Place node**:
There is a list: ["mirror_2", "lamp_1", "picture_7", "tool_6","toilet_8","sofa_11", "floor_12", "wall_13"].
**Question**: Please select one object that is most likely located near a sink. Always follow the format: Answer: <your answer>. Reasoning: <your Reasoning>.

(@Example LLM reply@)
**Answer**: mirror_2
**Reasoning**: Among the given options, the object most likely located near a sink is a "mirror." Mirrors are commonly found near sinks in bathrooms for personal grooming and hygiene activities.
\end{lstlisting}
\end{llmpromptbox}

\section{Controller details}
\label{app:controller_details}

We employ the ViNT~\cite{shah2023vint} GNM zero-shot as a cross-embodiment/environment image-goal visuomotor policy. We command ViNT to go to nearby objects by passing image crops of the object, extracted by the object detector and stored in the \graphshort{}. The \textit{Pathfinder} in the \textit{Reasoner} module issues subgoals that are within or connected to the current Place node, making it likely that the object subgoal is within the sensor's field-of-view. Additionally, we try to provide an updated, recent image of the subgoal to ViNT, by using object features to associate the image crop from the \graphshort{} with detections in the current RGB observation: if a successful association is made, we use the detected bounding box of the newly associated object as the subgoal.

We find empirically that ViNT is capable of local navigation and obstacle avoidance in this setting. ViNT is designed to output a temporal distance metric, which is used to check closeness to a subgoal. We found that the values this metric takes can vary with scene and embodiment, making it challenging to directly threshold on it as a termination condition for navigation. Instead, we find a heuristic adaptive thresholding approach based on this metric to be effective. Specifically, we set a threshold on the amount of positive change in temporal distance from the minimum recorded temporal distance to the subgoal, triggering a halt if change exceeds the threshold. This is based on the observation that while the mean values of the temporal distance metric vary across scene and embodiment, they exhibit consistent trends. In particular, the values decrease gradually until near the subgoal, after which they increase sharply, likely due to significant appearance change in the target subgoal stemming from the closer viewpoint.

\section{Simulation experiment setup details}
\label{app:sim_exp_details}

\subsection{Datasets}
We elaborate on the datasets chosen and evaluation metrics used in our simulation experiments. The selected \gibson{}~\cite{xiazamirhe2018gibsonenv} and \hmfull{}~\cite{ramakrishnan2021hm3d} datasets contain photorealistic 3D reconstructions of indoor environments. Specifically for \gibson{}, we use the validation set of the tiny split, comprising 1000 episodes evenly split over 5 household scenes, where each scene has 12-28 rooms across 3 floors. For \hm{}, we use 400 episodes from the validation set, sampled evenly across the 20 scenes in it. This doubles the size of the test set compared to the experimental setup of \citet{shah2023lfg}. Each scene in \hm{} contains between 4-19 rooms spread across 1-3 floors.

\subsection{Metrics}
We use 3 of the metrics used to benchmark submissions to the Habitat Challenge for ObjectNav~\cite{habitatchallenge2022}:
\begin{itemize}
    \item \textbf{Success rate (SR)}: The ratio of episodes where the agent succeeds. The Habitat evaluation setup determines the agent to have succeeded if it reaches a position within 1.0m from the goal instance, and the goal instance is within the robot's field-of-view for some orientations at that position.
    \item \textbf{Success-weighted path length (SPL)}: Path length travelled by agent over the episode weighted by binary success indicator variable. Intuitively measures the efficiency of the agent's path compared to the optimal path. Given by $$\frac{1}{N}\sum\limits_{i=1}^{N}S_i\frac{l_i}{\text{max}(p_i, l_i)},$$ where $l_i$ is the length of the shortest path between start position and goal instance in the current episode, $p_i$ is the length of the path driven by the agent over the entirety of the current episode, $S_i$ is the indicator variable for success.
    \item \textbf{Distance-to-goal (DTG)}: The agent's distance in metres from the goal instance at the end of the episode.
\end{itemize}

\subsection{Baselines}

We provide additional implementation details for the baselines described in \autoref{sec:sim_exp_setup}. We implement the LLM baselines, LGX-GT~\cite{dorbala2024lgx} and LFG-GT~\cite{shah2023lfg} by adapting their publicly available source code for the ObjectNav setup. We implement a Fast Marching Method (FMM)~\cite{sethian1996fmm} controller based on the code by \citet{chaplot2020semexp}, and standardise all LLM baselines (LGX-GT, LFG-GT) and variants of our approach (\oursprgt{}, \ourspr{}) to use this as a local planner, so as to compare their performance on the basis of differences in scene representation and reasoning.

The FMM-based controller takes in depth images and ground-truth pose from the simulator, and outputs linear/angular velocities. Similar to \citet{chaplot2020semexp}, our implementation builds a 2D obstacle map from the input observations, computes a shortest path over it using FMM, then converts this path into a sequence of control actions to execute. Our implementation additionally employs a heuristic recovery policy that attempts to rotate and perturb the robot in-place when it gets stuck or cannot find a valid path. 

For LGX-GT and LFG-GT, goals are specified to the controller as $(x, y)$ coordinates. For the \navsys{} variants \oursprgt{} and \ourspr{}, we adapt the controller to implement \textsc{MoveToObject} by accepting bounding boxes of detected objects as goals. To handle this, the FMM controller projects each bounding box detected from RGB inputs onto the corresponding depth image, computes the centroid of the 3D points within the box, then sets the centroid's $(x, y)$ coordinates as the goal. 

For the remaining ObjectNav approaches (SemExp~\cite{chaplot2020semexp}, PONI~\cite{ramakrishnan2022poni}, FBE~\cite{yamauchi1997fbe}, SemUtil~\cite{chen2023semutil}) we directly report their results on the \gibson{} dataset. We do not modify their controllers since our goal is to compare overall system performance, rather than ablating on sub-components like representations or reasoning.

\subsection{\graphshort{} specification}
\label{app:sim_exp_osg_spec}
We use the following \graphshort{} specification for all evaluations in simulation that make use of the full \engineshort{} module - including our ObjectNav evaluations, \graphshort{} building tests \textbf{(Q4)} and open-vocabulary goal generalisation tests \textbf{(Q5)}.

\begin{specbox}[specstyle, label=listing:sim_homes]{Simulation household environments}
\begin{lstlisting}[language=json]
"floor": {
    "layer_type": "Region Abstraction",
    "layer_id": 4,
    "contains": ["room"],
    "connects to": ["stairs"]
},
"room": {
    "layer_type": "Place",
    "layer_id": 3,
    "contains": ["object"],
    "connects to": ["entrance", "room", "stairs"]
},
"stairs": {
    "layer_type": "Connector",
    "layer_id": 2,
    "is near": ["object"],
    "connects to": ["floor", "room"]
},
"entrance": {
    "layer_type": "Connector",
    "layer_id": 2,
    "is near": ["object"],
    "connects to": ["room"]
},
"object": {
    "layer_id": 1,
},
"state": ["room"]
\end{lstlisting}
\end{specbox}

\subsection{Object feature evaluation}

We provide details on the experimental setup used to evaluate the accuracy of using object features for data association (\textbf{Q3} from \autoref{sec:sim_exp_setup}). We created a dataset of objects with associated object features, from views sampled across 10 \hm{} scenes. We selected the 10 largest environments from the \hm{} validation set, then selected all Place nodes from them that lie in the set \{living-room, dining-room, bedroom, bathroom, kitchen\}, then finally sampled Object/Connector nodes contained/connected to each Place node. This gives us a dataset comprising $\sim60$ rooms and $\sim150$ objects.

For each Place, Object and Connector node, we extracted their corresponding \textit{object features} from two different viewpoints. We employ the same object feature extraction pipeline described in the \engineshort{} (\autoref{sec:approach_mapper}). Briefly, the pipeline extracts objects from a view using an open-set object detector (GroundingDINO~\cite{liu2023grounding}), thresholds on pixel distance to obtain nearby Objects/Connectors, then queries a VQA model (BLIP2~\cite{li2023blip2}) to describe the colour, material and texture of each nearby Object/Connector. The aggregated list of descriptions of each nearby Object/Connector node constitutes the object features for a given Object/Connector. For each Place node, the object features are extended to include all Object/Connector nodes that are contained/connected to the Place node. The dataset consists of tuples $(i, v, t_{i,v}, L_{i,v}=[s_1,...,s_k])$, where $i$ is the node index, $v\in\{0,1\}$ is the viewpoint index, $t_{i,v}$ is the node type, and $L_{i,v}$ is the object features, where each $s_j$ is a text string describing a neighbouring Object/Connector node.

This dataset allows us to test the efficacy of object features as descriptors, and the accuracy of using them to recognise and distinguish Objects/Connectors/Places. For each node type, we computed the accuracy of \textit{recognising} the same Object/Connector/Place from two different viewpoints, \ie{} evaluating accuracy of correctly predicting that all pairs $\{(i, 0, t_{i,0}, L_{i,0}), (i, 1, t_{i, 1}, L_{i, 1})\}$ are the same node. We also computed the accuracy of using object features to \textit{distinguish} different Object/Connector/Place nodes, \ie{} evaluating accuracy of predicting that all pairs $\{(i,v,t_{i,v},L_{i,v}), (j,v,t_{i,v},L_{i,v})\}$ where $i\ne j$, are not the same node.

\subsection{\graphshort{} quality evaluation}
\label{app:osg_quality_eval}

We evaluate the quality of \graphshort{}s constructed for diverse environments with the \engineshort{}. To do so, we mapped 5 selected \hm{} environments with the \engineshort{}. We ran the \navsys{} system, but manually selected object subgoals for the agent to navigate to instead of using LLM planning, to enforce complete coverage of the environment. All environments are multi-storey, with four environments having at least two floors. Each of the environments has roughly 10 rooms. To quantify the accuracy, we manually annotated scene graphs of the environments, and computed the precision and recall of the \engineshort{}'s node/edge prediction with respect to our manually annotated graphs. We provide qualitative examples of the scene graph mapping below in \autoref{app:scene_graph_results}, and also show online construction of the \graphshort{} in our supplementary video.

\section{Real-world experiment setup details}
\label{app:real_exp_details}

Our tests are conducted tests in an open-plan apartment environment comprising a living room, dining room and kitchen. The \graphshort{} specification used is:

\begin{specbox}[specstyle, label=listing:rw_rls]{Real-world open-plan apartment}
\begin{lstlisting}[language=json]
"room": {
    "layer_type": "Place",
    "layer_id": 3,
    "contains": ["object"],
    "connects to": ["entrance", "room"]
},
"entrance": {
    "layer_type": "Connector",
    "layer_id": 2,
    "is near": ["object"],
    "connects to": ["room"]
},
"object": {
    "layer_id": 1,
},
"state": ["room"]
\end{lstlisting}
\end{specbox}

\section{Ablations over choice of foundation models}
\label{app:fm_ablations}

\begin{wraptable}{r}{0.65\textwidth}
    \centering
    \caption{\textbf{Ablation study on \ourspr{}'s ObjectNav performance with different LLMs.} Results are based on 100 episodes, with 5 episodes sampled from each scene in the \hm{} validation set.}
    \label{tab_app:fm_ablations}
    \small
        \begin{tabular}{@{\extracolsep{4pt}}cccc@{}}
            \toprule
            \textbf{LLM} & \textbf{SR} ($\uparrow$) & \textbf{SPL} ($\uparrow$) & \textbf{DTG} ($\downarrow$) \\
            \midrule
            Meta Llama 2 (7B)~\cite{touvron2023llama} & 0.349 & 0.095 & 4.190  \\
            Meta Llama 2 (13B)~\cite{touvron2023llama} & 0.538 & 0.124 & 2.867\\
            OpenAI GPT-3.5 &\textbf{0.693} & \textbf{0.282} & \textbf{2.385} \\
            \bottomrule
        \end{tabular}
\end{wraptable}

We ablate on the model choices used in \navsys{}. While \navsys{} employs 4 types of models: an LLM, a GNM and VFMs for VQA and open-set object detection, we focus on ablating over choices of LLM. In particular, we do not ablate over different GNMs since there \citet{shah2023vint} provide extensive performance evaluation, and we do not ablate over open-set object detection VFMs because there are few such models.

LLMs are essential to \navsys{} for semantic reasoning and belief updating, and for open-world knowledge about human environments. GPT-3.5's key limitation is its monetarily and computationally costly remote inference in the cloud. We explore the performance envelope of \navsys{} when using smaller models - \ie{} variants of Meta's LLaMa 2~\cite{touvron2023llama} - which are more efficient and enable local inference. Specifically, we compare GPT-3.5 against the 7B and 13B variants in \autoref{tab_app:fm_ablations}, which we run locally on a single Nvidia A5000 GPU. We find that performance is halved when using the 7B LLaMa 2 variant, owing to \textbf{(1)} its poor performance in predicting stop tokens, causing infinite recursion; \textbf{(2)} its tendency to reply to the prompts used in our state estimator with ``cannot decide''; \textbf{(3)} mistakes in classifying detections into Objects, Connectors and Places when updating the \graphshort{}. These challenges are mitigated with the 13B variant. While performance remains markedly poorer than GPT-3.5, we believe these smaller models offer a reasonable tradeoff, and consider ongoing work in more efficient LLMs to hold promise for handling \navsys{}'s current limitations.

\section{Additional results with open-set object queries}
\label{app:open_vocab}

\begin{wraptable}{r}{0.5\textwidth}
    \centering
    \caption{\textbf{Performance of \ourspr{} on rare object queries.}}
    \label{tab_app:rare_goals}
    \footnotesize
        \begin{tabular}{@{\extracolsep{4pt}}cccc@{}}
            \toprule
            \textbf{Object goal} & \textbf{SR} ($\uparrow$) & \textbf{SPL} ($\uparrow$) & \textbf{DTG} ($\downarrow$) \\
            \midrule
            \texttt{settee} & 0.8 & 0.285  & 0.894 \\
            \texttt{nightstand} & 0.7 &  0.287 & 1.699  \\
            \bottomrule
        \end{tabular}
\end{wraptable}

As described in \textbf{Q5} (\autoref{sec:sim_exp_setup}), we evaluate the ability to handle open-vocabulary goals by considering \textit{rare queries} and \textit{compositional queries}. As standard ObjectNav evaluation setups only consider 6 goal classes, we design new navigation episodes in \hm{} scenes with open-vocabulary goals. 

We define two rare object nouns - ``settee'' and ``nightstand'' - and build navigation episodes for each. We take reference from LVIS~\cite{gupta2019lvis}, which assembles a large-scale dataset and ranks the most commonly occurring objects. We select ``settee'' as it is an uncommon synonym not found in LVIS, though it refers to couches, which are a common object in household environments. We also select ``nightstand'' as it is an uncommon object completely absent from LVIS. \autoref{tab_app:rare_goals} shows that \navsys{} is able to effectively locate rare objects, owing to the strong open-vocabulary abilities of its VFMs and LLM.

We define several different types of compositional queries, by combining an object noun with one or more phrases describing properties like colour, material and pattern. While the individual nouns and adjectives may be common, the resulting compositional queries are uncommon, and unlikely to occur frequently (or at all) in the foundation models' training data. Examples of the nouns, adjectives and various composed queries are given in \autoref{tab_app:compositional_goals}. Our designed navigation episodes require \navsys{} to search for object instances matching these composed queries, and also to disambiguate the goal object instance from other objects with a similar class on the basis of the description: \eg{} distinguishing the target ``leopard-print chair'' from other chairs in the home. We find that \navsys{} shows strong performance, though we note that success can drop as descriptors are added and requirements on object appearance become more complex, making them susceptible to errors in perception.

\begin{table}[tbp]
\caption{\textbf{Success rate of \ourspr{} in handling open-vocabulary compositional queries.} In each query, an object goal is composed with one or more phrases describing its colour, material or pattern. E.g. a ``red, floral bed'' (C+P with ``bed'' as the object goal).}
\scriptsize
    \begin{center}
        \begin{tabular}{@{\extracolsep{4pt}}ccccccccccc@{}}
            \toprule
            \textbf{}&\textbf{}&\textbf{}&\textbf{}&\multicolumn{7}{c}{Success rate (over 10 trials each)}\\
            \cmidrule{5-11}
            \textbf{Object goal} & \textbf{Colour (C)} & \textbf{Material (M)} & \textbf{Pattern (P)} & \textbf{C} & \textbf{M} & \textbf{P} & \textbf{C+M} & \textbf{C+P} & \textbf{M+P} & \textbf{C+M+P}\\
            \midrule
            \texttt{chair} & \textit{black and white} & \textit{fabric} & \textit{leopard-print} & 0.8 & 0.6 & 0.8 & 0.6 & 0.6 & 0.6 & 0.6 \\
            \texttt{bed} & \textit{red} & \textit{cotton} & \textit{floral} & 0.8 & 0.8 & 0.8 & 0.6 & 0.6 & 0.6 & 0.4\\
            \bottomrule
        \end{tabular}
    \label{tab_app:compositional_goals}
    \end{center}
\end{table}

\section{Additional \graphshort{} construction results}
\label{app:scene_graph_results}

\autoref{app:osg_quality_eval} provides details on how we evaluate \graphshort{} construction quality over diverse \hm{} environments. In this section, we focus on providing qualitative examples of the \graphshort{}s built during this evaluation, as well as examples of \graphshort{}s built in real-world settings.

\subsection{Simulation: Examples in household environments}

We provide qualitative examples of \graphshort{}s built over a diverse range of environments, with a teleoperated agent. In these examples, we run the full \engineshort{} pipeline to perform state estimation and integration of observations into the \graphshort{}. We show in simulation that \graphshort{}s can be built with reliable accuracy over a variety of different multi-storey homes. We also show that the \engineshort{} is capable of handling significant variety in \graphshort{} specification, by mapping environments ranging from a supermarket (simulation) to a mall and an office (real-world).

\begin{figure}
    \centering
    \includegraphics[width=0.95\textwidth]{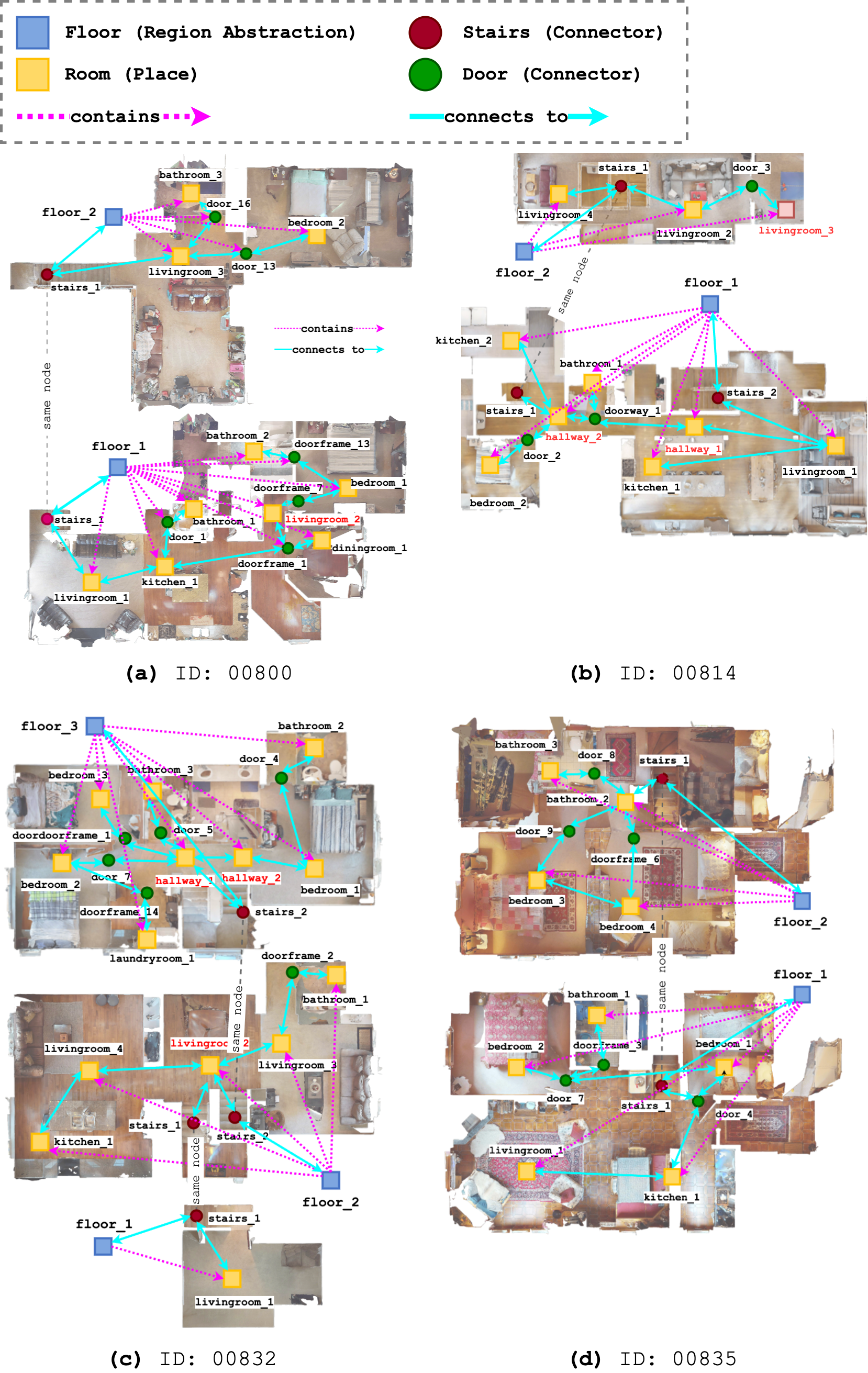}
    \caption{\textbf{\graphshort{} construction results in 4 \hm{} scenes.} The built \graphshort{}s are largely accurate and representative of the scenes, though there are some errors highlighted with \textcolor{red}{red labels}.}
    \label{fig_app:hm3d_sg_examples}
\end{figure}

\autoref{fig_app:hm3d_sg_examples} provides examples of \graphshort{}s built in 4 different multi-storey scenes from \hm{}. We use the same \graphshort{} specification detailed in \autoref{app:sim_exp_osg_spec}. Qualitatively, \engineshort{} correctly identifies most of the room (Place), floor (Region Abstraction) nodes and edges. Its main failure modes occur in over-segmenting hallways (see 00814, 00832) and misclassifying hallways as other rooms (see 00800, 00832, where hallways are misclassified as living-rooms). The latter occurs especially when significant area in the living-room is visible from along much of the hallway, leading our VQA model to misidentify the hallway as a living-room.

\subsection{Simulation: Examples in supermarket}
\label{app:scene_graph_supermarket}

\begin{specbox}[specstyle, label=listing:osg_spec_supermarket]{Supermarket}
\begin{lstlisting}[language=json]
"aisle": {
    "layer_type": "Place",
    "layer_id": 3,
    "contains": ["object"],
    "connects to": ["aisle"]
},
"object": {
    "layer_id": 1,
},
"state": ["aisle"]
\end{lstlisting}
\end{specbox}

\begin{figure}
    \centering
    \includegraphics[width=\textwidth]{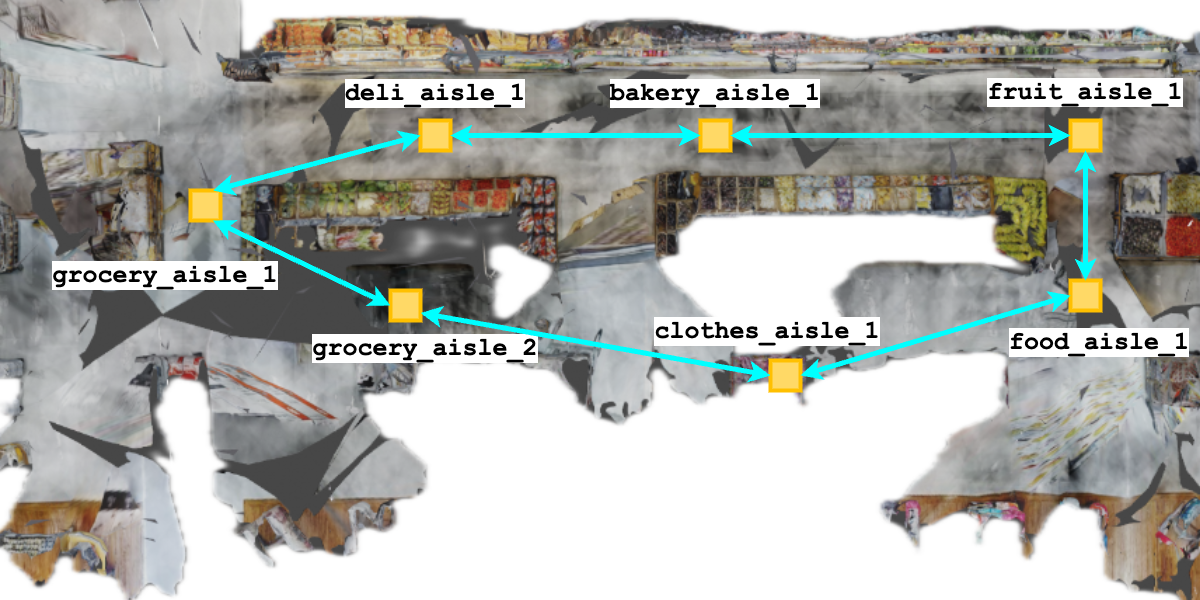}
    \caption{\textbf{\graphshort{} of a simulated supermarket from Gibson (\textit{Gratz}).} The \engineshort{} is able to use the spatial concepts described in \autoref{listing:osg_spec_supermarket} to recognise and produce semantically meaningful labels for different aisles (Places) in the supermarket, and can localise and build a coherent \graphshort{} of this different scene.}
    \label{fig_app:gratz_sg_example}
\end{figure}

We demonstrate the \engineshort{}'s versatility in handling varied \graphshort{} specifications containing open-vocabulary spatial concepts. We do so by specifying the supermarket \graphshort{} specification shown in \autoref{listing:osg_spec_supermarket}, which uses the more appropriate abstraction of ``aisles'' as Place nodes. Using this specification, we build the \graphshort{} of Gibson's \textit{Gratz} supermarket scene (\autoref{fig_app:gratz_sg_example}), and show that \navsys{} is able to accurately capture the ``aisle'' abstraction.

\subsection{Real-world: Examples in mall and office environments}

\begin{figure}
    \centering
    \includegraphics[width=\textwidth]{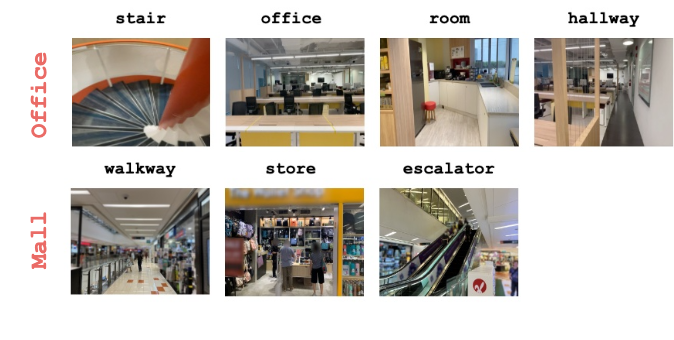}
    \caption{\textbf{Spatial concepts from Mall/Office \graphshort{} specs.} Full \graphshort{} specifications using these concepts are given in \autoref{listing:osg_spec_office} and \autoref{listing:osg_spec_mall}.}
    \label{fig_app:sg_rw_elements}
\end{figure}

We build on the simulation tests in \autoref{app:scene_graph_supermarket} with real-world tests in mall and office environments to further underscore the \engineshort{}'s versatility in handling diverse, open-vocabulary \graphshort{} specifications. \autoref{fig_app:sg_rw_elements} highlights the variety of spatial concepts used in the \graphshort{} specifications. We capture the Office environment with spatial concepts similar, but also finer-grained than those used in household environments: \ie{} we more finely distinguish between rooms, offices and hallways. The mall environment uses a completely distinct set of spatial concepts, with Places in the form of stores and walkways, and Connectors like escalators. We describe the \graphshort{} specifications in detail in \autoref{listing:osg_spec_office} and \autoref{listing:osg_spec_mall}.

\begin{specbox}[specstyle, label=listing:osg_spec_office]{Office}
\begin{lstlisting}[language=json]
"floor": {
    "layer_type": "Region Abstraction",
    "layer_id": 4,
    "contains": ["hallway", "office", "room"],
    "connects to": ["stairs"]
},
"hallway": {
    "layer_type": "Place",
    "layer_id": 3,
    "contains": ["object"],
    "connects to": ["hallway", "office", "room", "entrance"]
},
"office": {
    "layer_type": "Place",
    "layer_id": 3,
    "is near": ["object"],
    "connects to": ["hallway", "office", "room", "entrance"]
},
"room": {
    "layer_type": "Place",
    "layer_id": 3,
    "is near": ["object"],
    "connects to": ["hallway", "office", "room", "entrance"]
},
"stair": {
    "layer_type": "Connector",
    "layer_id": 2,
    "is near": ["object"],
    "connects to": ["floor"]
},
"entrance": {
    "layer_type": "Connector",
    "layer_id": 2,
    "is near": ["object"],
    "connects to": ["hallway", "office", "room"]
},
"object": {
    "layer_id": 1,
},
"state": ["hallway", "office", "room"]
\end{lstlisting}
\end{specbox}

\begin{specbox}[specstyle, label=listing:osg_spec_mall]{Mall}
\begin{lstlisting}[language=json]
"floor": {
    "layer_type": "Region Abstraction",
    "layer_id": 4,
    "contains": ["walkway", "store"],
    "connects to": ["escalator"]
},
"walkway": {
    "layer_type": "Place",
    "layer_id": 3,
    "contains": ["object"],
    "connects to": ["walkway", "store", "escalator"]
},
"store": {
    "layer_type": "Place",
    "layer_id": 3,
    "is near": ["object"],
    "connects to": ["store", "walkway"]
},
"escalator": {
    "layer_type": "Connector",
    "layer_id": 2,
    "is near": ["object"],
    "connects to": ["floor", "walkway"]
},
"object": {
    "layer_id": 1,
},
"state": ["walkway", "store"]
\end{lstlisting}
\end{specbox}

\begin{figure}
    \centering
    \includegraphics[width=\textwidth]{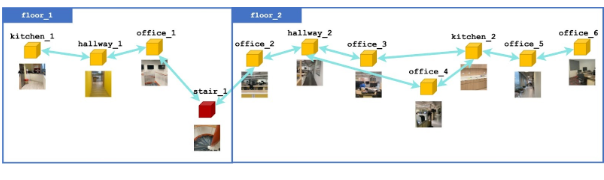}
    \caption{\textbf{\graphshort{} of real world Office environment.}}
    \label{fig_app:sg_rw_office}
\end{figure}

\begin{figure}
    \centering
    \includegraphics[width=\textwidth]{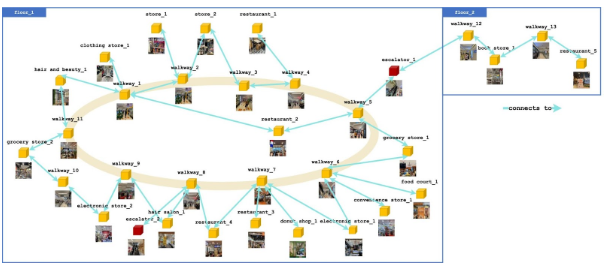}
    \caption{\textbf{\graphshort{} of real-world Mall environment.}}
    \label{fig_app:sg_rw_mall}
\end{figure}

We show the results of \graphshort{} construction for the Office and Mall respectively in \autoref{fig_app:sg_rw_office} and \autoref{fig_app:sg_rw_mall}. We find that \engineshort{} is overall able to correctly interpret the spatial concepts in the \graphshort{} specifications and label the different locations in the Office and Mall according to these specified concepts. The main failure mode encountered is that the \engineshort{} tends to over-segment long continuous Place regions - \eg{} a single walkway in the real-world mall tends to be split into multiple walkways in the \graphshort{}. This is likely due to the dearth of spatial, geometric information in the \navsys{} system. We consider the possibility of incorporating some limited, local geometric information into the system a promising research direction, possibly employing recent foundation models for geometric scene understanding such as DepthAnything.

\section{Additional Discussion}
\label{app:add_discussion}

Overall, our experiments highlight the feasibility and promise of composing a system for OWON from foundation models and \graphshort{}s. However, we also provide additional discussion about the limitations of our approach, along with future directions.

Firstly, \navsys{} relies computationally and monetarily expensive querying of LLMs, as described in \autoref{sec:conclusion}. Our initial experiments using smaller models in a local inference setting are promising (\autoref{app:fm_ablations}), as they show that significantly smaller models than GPT-3.5 can reach reasonable levels of performance in a zero-shot setting comparable with older ObjectNav approaches trained on curated datasets. We believe that ongoing work on building smaller language models, and making such models more effective at reasoning~\cite{mitra2023orca} holds promise for making \navsys{} more practical in everyday robotics.

Secondly, \navsys{} does not handle uncertainty, especially perceptual and state estimation uncertainty. The system can be made more robust by representing uncertainty and considering multiple hypotheses in \graphshort{} mapping. A key difficulty in doing this lies in accurately quantifying the uncertainty of LLM responses, which is crucial given our extensive use of LLMs. We plan to explore possible solutions offered by recent works~\cite{ren2023knowno, ren2024explore}, which apply tools like conformal prediction to calibrating confidence in embodied planning and question-answering with LLMs and VLMs.

Thirdly, \navsys{} currently does not use any metric information. As the structure of a 3D environment can be ambiguous from a 2D image, this can lead to inaccurate \graphshort{} mapping results. This occasionally leads to failures such as inaccurate classification of Places when significant amounts of Objects from other Places are visible within the agent's field-of-view. This occurs since the \engineshort{} is not explicitly considering the 3D geometry of the environment in its classification. While the non-metric \engineshort{} still proves to be fairly effective for zero-shot ObjectNav, we plan to augment it with local geometric information in more specific settings (\eg{} where the robot embodiment is known), allowing \graphshort{}s to more accurately understand and represent the scene. Moreover, recent works on foundation models for geometric scene understanding like DepthAnything potentially offer a generalisable means toward incorporating metric and geometry information into our system.

Finally, \navsys{} can enable lifelong learning through the ``promptability'' of its \graphshort{} representation and its subsystems. Lifelong learning refers to learning or adapting policies and representations to improve task performance or adjust to new environments over the robot's lifetime. Most scene graph approaches have fixed structures that cannot be adapted without fundamental modifications to the robot system, rendering them unable to accurately represent the scene if their structures do not match the environment. In contrast, \graphshort{}s can be easily ``prompted'' to adopt a particular structure suiting the scene. The ability to dynamically adapt the structure with different \graphshort{} specifications paves the way for online adaptation of \graphshort{}s to handle new environments. We intend to explore online, automatic inference and updating of the \graphshort{} specification as the robot navigates, potentially allowing \navsys{} to autonomously adapt to representing new scenes. This can enable \graphshort{} mapping and hence \navsys{} to scale to large, heterogeneous environments, which prior scene graph approaches might struggle to represent. Separately, most of \navsys{}'s subsystems depend on LLMs that can be prompted online to improve performance. In particular, online prompting of historical context can improve planning and exception handling with LLMs, and LLMs can also be prompted with newly acquired scene information to improve semantic reasoning.

\end{document}